\documentclass{article}

\usepackage{microtype}
\usepackage{graphicx}
\usepackage{subcaption}
\usepackage{booktabs} 

\usepackage{hyperref}

\usepackage[preprint]{icml2026}

\usepackage{amsmath}
\usepackage{amssymb}
\usepackage{mathtools}
\usepackage{amsthm}

\usepackage[capitalize,noabbrev]{cleveref}

\theoremstyle{plain}

\theoremstyle{definition}

\theoremstyle{remark}

\usepackage[textsize=tiny]{todonotes}

\usepackage{graphicx}

\usepackage{hyperref}
\usepackage{url}

\usepackage{graphicx}%
\usepackage{multirow}%
\usepackage{amsmath,amssymb,amsfonts}%
\usepackage{amsthm}%
\usepackage{mathrsfs}%
\usepackage[title]{appendix}%
\usepackage{xcolor}%
\usepackage{textcomp}%
\usepackage{manyfoot}%
\usepackage{booktabs}%
\usepackage{algorithm}%
\usepackage{listings}%
\usepackage{float}
\usepackage{placeins}
\usepackage[table]{xcolor} 
\usepackage{colortbl} 
\usepackage{verbatim}
\usepackage{tikz}

\definecolor{tablegreen}{rgb}{0.91, 0.94, 0.97}
\definecolor{LightCyan}{rgb}{0.80, 1, 1}
\definecolor{LightPink}{rgb}{1, 0.89, 0.88}

\newcommand{\std}[1]{{\scriptsize(#1)}}

\newcommand{\highlight}[2]{\tikz[baseline]{\node[fill=#1, anchor=base, minimum width=2em, inner sep=1pt, outer sep=0.5pt, rounded corners] {#2};}}

\icmltitlerunning{How Creative Are Large Language Models in Generating Molecules?}

\begin{document}

\twocolumn[
  \icmltitle{How Creative Are Large Language Models in Generating Molecules?}



  \icmlsetsymbol{equal}{*}

  \begin{icmlauthorlist}
    \icmlauthor{Wen Tao}{1}
    \icmlauthor{Yiwei Wang}{2}
    \icmlauthor{Peng Zhou}{3}
    \icmlauthor{Bryan Hooi}{4}
    \icmlauthor{Wanlong Fang}{1}\\
    \icmlauthor{Tianle Zhang}{1}
    \icmlauthor{Xiao Luo}{3}
    \icmlauthor{Yuansheng Liu}{3}
    \icmlauthor{Alvin Chan}{1}
  \end{icmlauthorlist}

  \icmlaffiliation{1}{Nanyang Technological University}
  \icmlaffiliation{2}{University of California, Merced}
  \icmlaffiliation{3}{Hunan University}
  \icmlaffiliation{4}{National University of Singapore}

  \icmlcorrespondingauthor{Wen Tao}{taowen228@gmail.com}
  \icmlcorrespondingauthor{Alvin Chan}{guoweialvin.chan@ntu.edu.sg}

  \icmlkeywords{Machine Learning, ICML}

  \vskip 0.3in
]



\printAffiliationsAndNotice{}  

\begin{abstract}
Molecule generation requires satisfying multiple chemical and biological constraints while searching a large and structured chemical space. This makes it a non-binary problem, where effective models must identify non-obvious solutions under constraints while maintaining exploration to improve success by escaping local optima. From this perspective, creativity is a functional requirement in molecular generation rather than an aesthetic notion. Large language models (LLMs) can generate molecular representations directly from natural language prompts, but it remains unclear what type of creativity they exhibit in this setting and how it should be evaluated. In this work, we study the creative behavior of LLMs in molecular generation through a systematic empirical evaluation across physicochemical, ADMET, and biological activity tasks. We characterize creativity along two complementary dimensions, \emph{convergent creativity} and \emph{divergent creativity}, and analyze how different factors shape these behaviors. Our results indicate that LLMs exhibit distinct patterns of creative behavior in molecule generation, such as an \emph{increase} in constraint satisfaction when additional constraints are imposed. Overall, our work is the first to reframe the abilities required for molecule generation as creativity, providing a systematic understanding of creativity in LLM-based molecular generation and clarifying the appropriate use of LLMs in molecular discovery pipelines.
\end{abstract}

\section{Introduction}

Designing new functional molecules aims to identify chemical structures that satisfy desired properties, such as selective inhibition of a disease target, while also meeting additional constraints related to stability, synthesizability, and safety~\cite{gao2022sample}. As a result, molecule generation is not a binary problem with strictly correct or incorrect answers. For a given request, there often exist many chemically valid solutions, and what distinguishes strong models is whether they can discover non-obvious structures that satisfy the constraints, rather than reproducing close variants of known examples. Given the high failure rate in drug discovery~\cite{dimasi2010trends}, generating diverse designs is important for mitigating failure risk and navigating intellectual property restrictions.

\begin{figure}[t]
  \centering
    \includegraphics[width=\linewidth]{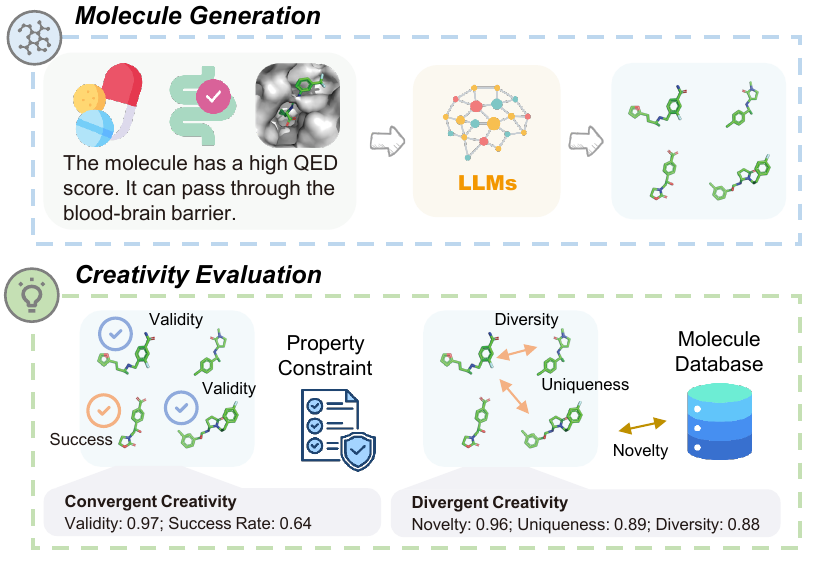}
  \caption{Overview of how LLM creativity in molecule generation is evaluated.
\textbf{Top:} LLMs generate molecules conditioned on constraints (physicochemical, ADMET, and biological activity) specified in the prompt.
\textbf{Bottom:} Creativity is operationalized along two complementary dimensions: convergent creativity (constraint satisfaction) and divergent creativity (chemical space exploration).}
  \label{fig:framework}
\end{figure}

From this perspective, molecule generation inherently involves creativity. In this paper, we treat the ability required for molecule generation as creativity in a functional sense. A creative molecular generator should produce chemically valid molecules that satisfy task constraints, while also exploring new regions of chemical space that may contain viable candidates. The central challenge lies in balancing constraint satisfaction with meaningful exploration, which makes creativity in molecule generation a scientific question with direct practical relevance rather than a subjective or aesthetic notion.

Given the importance of molecular design, substantial effort has been devoted to systematic approaches that avoid exhaustive search~\cite{cheng2021molecular}. Before the rise of large language models (LLMs), creativity in molecular generation was mainly explored through specialized generative models such as variational autoencoders~\cite{jin2018junction} and diffusion-based models~\cite{hoogeboom2022equivariant}. 

LLMs introduce a qualitatively different paradigm. Trained on large-scale text corpora that contain chemical knowledge, LLMs can generate molecular representations directly from natural language prompts, combine multiple constraints without retraining, and adapt their outputs through prompting or in-context examples~\cite{li2024empowering,wang2023grammar,tao2025make}. This flexibility has enabled molecule generation using language-only interfaces, raising the possibility that creativity in molecular design could be studied through general-purpose language models.

Despite these advances, it remains unclear what kind of creativity LLMs actually exhibit in molecule generation, and how such creativity should be measured. In this work, we conduct an empirical study across physicochemical, ADMET, and biological activity tasks to address the following questions:

\begin{itemize}

  \item \textbf{\textit{How can creativity in molecule generation be operationalized to reflect both task performance and exploration?}} 
  As shown in Figure~\ref{fig:framework}, we characterize creativity along two complementary dimensions, convergent creativity and divergent creativity, corresponding to constraint satisfaction and chemical space exploration. This operationalization allows us to systematically analyze the inherent trade-off between these two aspects in molecule generation.

  \item \textbf{\textit{How do different factors shape the creative behavior of LLMs in molecule generation?}} We analyze the effects of task type, numerical constraints, model size, constraint complexity, in-context learning, and sampling temperature on both convergent and divergent creativity. Our findings indicate that LLMs show distinct patterns of creative behavior in molecule generation. For example, \textit{adding molecular property constraints can increase the fraction of generated molecules that satisfy the specified constraints.}

  \item \textbf{\textit{What implications do the observed behaviors reveal about the current use of LLMs for molecular discovery?}} 
  Our work provides practical recommendations to guide AI researchers and chemists in using LLMs for molecule generation. These guidelines clarify when LLMs are effective, where their limitations lie, and how to configure them to balance constraint satisfaction and exploration. For example, \textit{ LLMs are effective when constraints are coarse-grained and primarily determined by global molecular properties.}
\end{itemize}

Overall, this work provides a systematic understanding of creativity in LLM-based molecule generation and clarifies the appropriate use of LLMs in molecular discovery pipelines.

\begin{table*}[!ht]
\centering
\caption{Evaluation framework for LLM creativity in molecule generation. Convergent metrics assess constraint satisfaction, divergent metrics assess exploration, and composite metrics balance both dimensions. $\operatorname{GM}(\cdot)$ denotes the geometric mean.}
\resizebox{\textwidth}{!}{
\begin{tabular}{@{}llll@{}}
\toprule
\textbf{Metric Category} & \textbf{Metric Name} & \textbf{Definition / Purpose} & \textbf{Role in Framework} \\ \midrule
Convergent & Validity & Are the molecules chemically real? & Foundational Check \\
 & Success Rate & Do they meet the prompt's constraints? & Constraint Check \\ \midrule
Divergent & Novelty & Are they different from known data? & Originality Check \\
 & Uniqueness & Are they different from each other? & Variety Check \\
 & Diversity & How structurally different is the batch? & Exploration Check \\ \midrule
 Composite & Convergent Creativity & $\operatorname{GM}(\text{Validity}, \text{Success Rate})$ & Convergent Creativity Score \\
  & Divergent Creativity & $\operatorname{GM}(\text{Novelty}, \text{Uniqueness}, \text{Diversity})$ & Divergent Creativity Score \\
  & Overall Creativity & $\operatorname{GM}(\text{Convergent Creativity}, \text{Divergent Creativity})$ & Overall Creativity Score \\ \midrule
Elite Subset & Fully Creative & \begin{tabular}[c]{@{}l@{}}Proportion simultaneously Novel, \\ Unique, \& Successful.\end{tabular} & The ``Gold Standard'' hit rate \\ \bottomrule
\end{tabular}
}
\label{tab:creativity_framework}
\end{table*}

\section{Operationalizing Creativity in Molecule Generation}

We propose a framework for evaluating creativity in LLM-based molecular generation. As illustrated in Figure~\ref{fig:framework}, the objective of molecular generation is discovery, which requires balancing constraint satisfaction with exploration of chemical space. Overly unconstrained generation produces chemically invalid or unusable molecules, while overly constrained generation collapses toward known structures and limits discovery potential.

\subsection{Two dimensions of creativity}

Drawing on established creativity theory~\citep{guilford1967creativity}, we characterize molecular generation creativity along two complementary dimensions. \textbf{Convergent creativity} measures whether generated molecules satisfy explicit task constraints and basic chemical rules, capturing the model's ability to produce valid, task-compliant outputs. \textbf{Divergent creativity} measures the extent to which a model explores
chemical space by producing structurally varied and novel molecules, capturing the model's capacity for discovery. These dimensions are distinct and must be evaluated separately to avoid conflating constraint satisfaction with exploratory behavior.

\subsection{Measuring convergent creativity}

Convergent creativity is quantified through two metrics. \textbf{Validity} evaluates whether generated strings correspond to chemically valid molecules according to standard cheminformatics rules. \textbf{Success Rate} measures the proportion of generated molecules that are both valid and satisfy all task-specific constraints (e.g., desired molecular properties). We aggregate these metrics into a single convergent creativity score using their geometric mean. The geometric mean ensures that both validity and constraint satisfaction must be reasonably high, preventing models from achieving high scores through extreme performance on only one dimension.

\subsection{Measuring divergent creativity}

Divergent creativity is quantified through three complementary metrics. \textbf{Novelty} evaluates whether generated molecules are absent from a reference dataset; following prior work~\cite{gao2022sample, li2024tomg}, we use ZINC 250K~\cite{sterling2015zinc} as the reference due to its pharmaceutical relevance, moderate size, and community adoption. We also include the DRD2, JNK3, and GSK3$\beta$ datasets~\cite{nakamura2025molecular, jin2020multi} in the reference dataset, as they are used for selecting in-context learning samples. \textbf{Uniqueness} measures the fraction of non-duplicate molecules within a generation batch, reflecting the model's ability to avoid repetitive outputs. \textbf{Diversity} quantifies average pairwise structural dissimilarity among generated molecules, measured using molecular fingerprints~\cite{rogers2010extended} and Tanimoto distance. We combine these metrics using their geometric mean as the measure for divergent creativity.

\subsection{Overall creativity}

To provide a unified measure capturing both dimensions of creativity, we define \textbf{Overall Creativity} as the geometric mean of convergent and divergent creativity. We assign equal importance to constraint satisfaction and exploratory diversity, similar to previous work in NLP~\cite{lu2025benchmarking}.

Finally, we define generated molecules that simultaneously satisfy task constraints, are novel with respect to the reference dataset, and are unique within the batch, as \textbf{Fully Creative}. The \textbf{Fully Creative \%} metric identifies the proportion of generations that satisfies both convergent and divergent creativity criteria.

Formal definitions and theoretical justification of all metrics are provided in Appendix~\ref{app:metrics}. Table~\ref{tab:creativity_framework} summarizes the complete evaluation framework. Here, $\operatorname{GM}(\cdot)$ denotes the geometric mean, defined as

\begin{equation}
\operatorname{GM}(x_1,\ldots,x_n) = \left(\prod_{i=1}^n x_i\right)^{1/n}.
\end{equation}

\section{Experimental Setup}

As shown in Figure~\ref{fig:framework} (Top), we evaluate an LLM's creativity in molecular generation by sampling from its prompt-conditioned distribution. 
The model is provided with a set of $N$ constraints
$\mathcal{C}_N = \{c_1, c_2, \dots, c_N\}$, where $N \ge 1$, and optionally $M$ example molecules
$\{\mathbf{y}^{(1)}, \dots, \mathbf{y}^{(M)}\}$ that satisfy these constraints, with $M \ge 0$ (for example, $M = 10$ in the few-shot setting).
When $M = 0$, the model operates in a zero-shot setting.
Conditioned on the provided constraints and examples, the model generates a SMILES representation for a molecule $\mathbf{y}$.

\begin{equation}
    \widehat{\mathbf{y}} \sim P_{\text{LLM}}\Big(\mathbf{y} \;\big|\; 
    \underbrace{\mathbf{c}^{(1)}, \dots, \mathbf{c}^{(N)}}_{\text{constraints}}, \; 
    \underbrace{\mathbf{y}^{(1)}, \dots, \mathbf{y}^{(M)}}_{\text{examples}}\Big).
\end{equation}

\noindent\textbf{Task description.} 
To systematically evaluate the capabilities of LLMs, we design eight single-constraint molecular generation tasks spanning three categories. The descriptions and prompts used for each task are summarized in Table~\ref{tab:tasks_prompts}.

\begin{table*}[htbp]
\centering
\caption{Summary of molecular generation tasks and corresponding prompts used in this work. [VALUE] denotes a real-valued number.}
\resizebox{\textwidth}{!}{
\begin{tabular}{llll}
\toprule
\textbf{Property Type} & \textbf{Task Definition} & \textbf{Prompt} \\
\midrule
Physicochemical & High quantitative estimate of drug-likeness (QED $\ge 0.6$) 
& The molecule has a high QED score. \\

 & Good synthetic accessibility (SA $\le 4$) 
& The molecule has good synthetic accessibility. \\

 & Specified hydrophilicity / hydrophobicity (LogP $\in \{-3,-1,1,3,5\}$) 
& The molecule has a LogP value of [VALUE]. \\ \midrule

ADMET & Blood--brain barrier permeability (BBB $\ge 0.5$) 
& The molecule can pass through the blood-brain barrier. \\

 & Human intestinal absorption (HIA $\ge 0.5$) 
& The molecule can be absorbed by the human intestine. \\ \midrule

Activity & High dopamine type 2 receptor affinity (DRD2 $\ge 0.5$) 
& The molecule can bind to DRD2. \\

 & High c-Jun N-terminal kinase-3 affinity (JNK3 $\ge 0.5$) 
& The molecule can bind to JNK3. \\

 & High glycogen synthase kinase-3 beta affinity (GSK3$\beta$ $\ge 0.5$) 
& The molecule can bind to GSK3$\beta$. \\

\bottomrule
\end{tabular}
}
\label{tab:tasks_prompts}
\end{table*}

\noindent\textbf{Evaluation.} To measure these properties, we use RDKit \cite{landrum2013rdkit} to calculate the LogP values of the generated molecules; TDC oracle functions \cite{huang2021therapeutics} to obtain predictive scores for target activity on DRD2, JNK3, and GSK3$\beta$, as well as QED and SA; and models from ADMET-AI \cite{swanson2024admet} to obtain predictive scores for BBB and HIA.

\noindent\textbf{Key details.} 
We conduct experiments across a diverse set of LLMs that vary in model size, series, and architecture, including LLaMA3-8B, LLaMA3-70B, LLaMA3.1-8B, LLaMA3.1-70B \cite{grattafiori2024llama}, GPT-3.5, GPT-4o-mini \cite{hurst2024gpt}, and DeepSeek-V3 \cite{liu2024deepseek}. The GPT series models are accessed via the OpenAI API. The other LLMs are accessed via the OpenRouter API. The default sampling temperature is set to 1 for all experiments. All reported results are averaged over five independent runs.

\section{Analyses of LLM Creativity}

\subsection{Is convergent creativity in conflict with divergent creativity?}

\textbf{There exists a trade-off between divergent and convergent creativity.} To examine the relationship between convergent and divergent creativity, we compute the Pearson correlation matrix among molecular generation creativity metrics, including validity, success rate (SR), novelty, uniqueness, and diversity. This analysis is based on experiments on physicochemical and ADMET tasks. We exclude activity tasks from this analysis. For activity tasks, the LLMs show limited task understanding, and the success rates are close to zero, which would lead to degenerate correlations that are not informative. The resulting Pearson correlation matrix is reported in Table~\ref{tab:pearson_corr_metrics}.

\begin{table}[ht]
\centering
\caption{Pearson correlation matrix of molecular generation creativity metrics, with positive correlations marked in pink and negative correlations marked in blue.}
\resizebox{0.5\textwidth}{!}{
\begin{tabular}{lccccc}
\toprule
\textbf{Metrics} & \textbf{Validity} & \textbf{SR} & \textbf{Novelty} & \textbf{Uniqueness} & \textbf{Diversity} \\
\midrule
Validity    & \highlight{LightPink}{1.000} & \highlight{LightPink}{0.986} & \highlight{LightCyan}{-0.747} & \highlight{LightCyan}{-0.889} & \highlight{LightCyan}{-0.799} \\
SR          & \highlight{LightPink}{0.986} & \highlight{LightPink}{1.000} & \highlight{LightCyan}{-0.704} & \highlight{LightCyan}{-0.805} & \highlight{LightCyan}{-0.721} \\
Novelty     & \highlight{LightCyan}{-0.747} & \highlight{LightCyan}{-0.704} & \highlight{LightPink}{1.000} & \highlight{LightPink}{0.857} & \highlight{LightPink}{0.962} \\
Uniqueness  & \highlight{LightCyan}{-0.889} & \highlight{LightCyan}{-0.805} & \highlight{LightPink}{0.857} & \highlight{LightPink}{1.000} & \highlight{LightPink}{0.956} \\
Diversity   & \highlight{LightCyan}{-0.799} & \highlight{LightCyan}{-0.721} & \highlight{LightPink}{0.962} & \highlight{LightPink}{0.956} & \highlight{LightPink}{1.000} \\
\bottomrule
\end{tabular}
}
\label{tab:pearson_corr_metrics}
\end{table}

\begin{figure*}[t]
  \centering
  \begin{subfigure}[b]{0.33\textwidth}
    \centering
    \includegraphics[width=\linewidth]{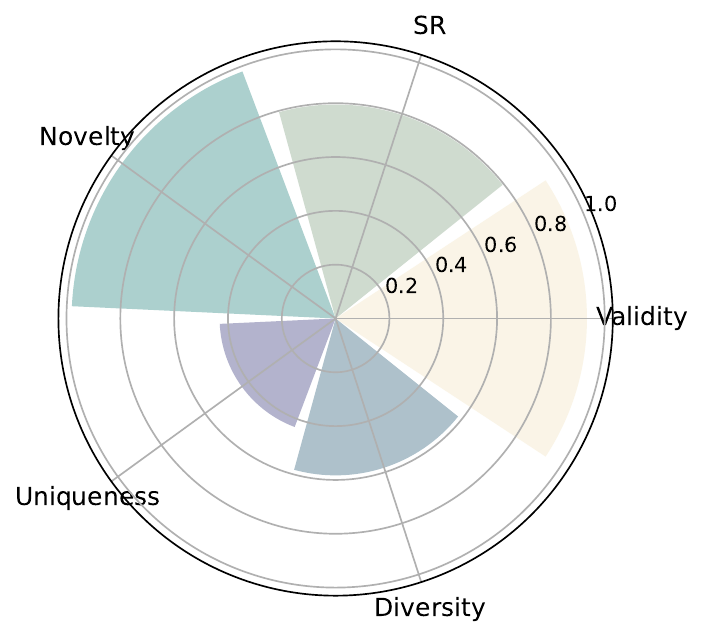}
    \caption{Physicochemical tasks}
    \label{fig:physicochemical}
  \end{subfigure}\hfill
  \begin{subfigure}[b]{0.33\textwidth}
    \centering
    \includegraphics[width=\linewidth]{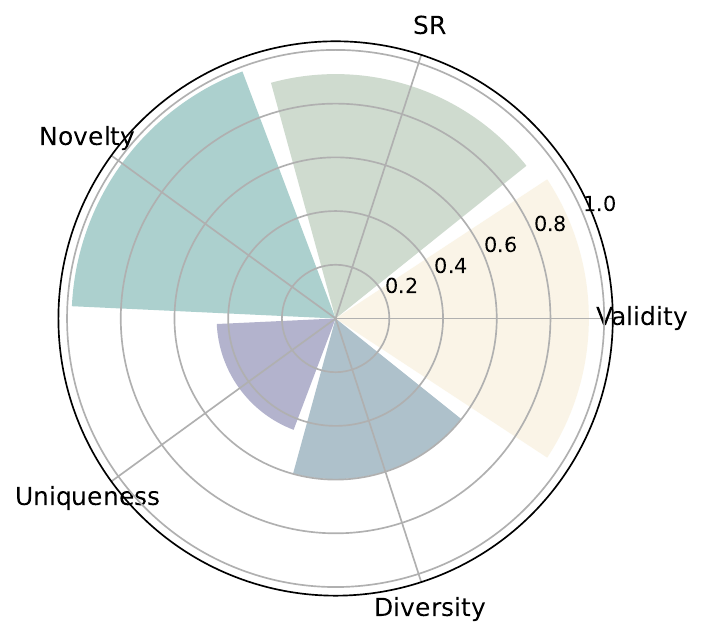}
    \caption{ADMET tasks}
    \label{fig:admet}
  \end{subfigure}\hfill
  \begin{subfigure}[b]{0.33\textwidth}
    \centering
    \includegraphics[width=\linewidth]{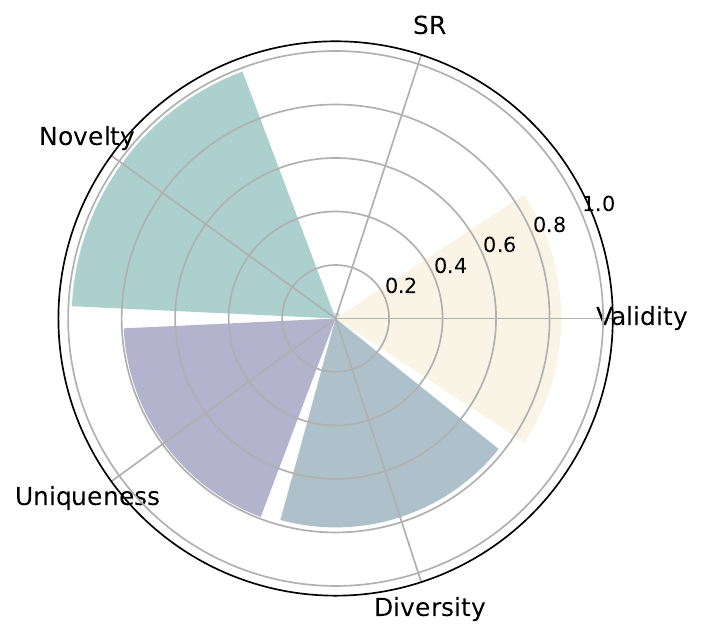}
    \caption{Activity tasks}
    \label{fig:target}
  \end{subfigure}
  \caption{Creativity profiles vary systematically by task type. Physicochemical and ADMET tasks show high convergent creativity but lower exploration, while activity tasks exhibit the opposite pattern.}
  \label{fig:average_results_on_three_type_tasks}
\end{figure*}

From the correlation analysis, several consistent patterns emerge. Validity and success rate exhibit an extremely strong positive correlation, indicating that they capture closely related aspects of goal satisfaction and constraint compliance, which we treat as convergent creativity. In contrast, both validity and success rate show strong negative correlations with novelty, uniqueness, and diversity. This indicates a systematic trade-off between convergent creativity and divergent creativity.

Within the divergent creativity metrics, novelty, uniqueness, and diversity are strongly positively correlated with each other. This suggests that, at the task level, these metrics reflect a shared tendency toward exploration rather than independent behaviors. Together, these results provide direct empirical evidence that improving convergent creativity is often accompanied by reduced divergent creativity, forming a fundamental tension in LLM-based molecular generation.

\subsection{Does LLM creativity differ across task types?}

\textbf{LLMs exhibit systematically different creativity profiles across task types.} 
Figure~\ref{fig:average_results_on_three_type_tasks} summarizes the average performance of all evaluated models on physicochemical, ADMET, and biological target activity tasks. On physicochemical and ADMET constraints (e.g., QED, SA, BBB, and HIA), general-purpose LLMs achieve surprisingly high success rates in a zero-shot setting. In contrast, for biological target-specific activity tasks such as DRD2, JNK3, and GSK3$\beta$, success rates consistently drop to nearly zero.

This sharp contrast indicates that LLMs can satisfy constraints (whether it is drug-like) that rely primarily on global molecular properties learned from large-scale text data, but fail on tasks (whether it is active against a specific protein) that require fine-grained protein--ligand interaction knowledge and task-specific structure--activity relationships.

Beyond success rate, task type also shapes the balance between convergent and divergent creativity. For physicochemical and ADMET tasks, higher success rates are accompanied by higher validity, indicating stable convergent behavior. However, these tasks exhibit lower uniqueness and diversity compared to target-specific activity tasks. In contrast, activity tasks show much higher divergent creativity despite near-zero success rates, reflecting broad exploration without effective constraint satisfaction.

Overall, these observations show that task type directly influences not only whether LLMs succeed, but also how convergent and divergent creativity are expressed.

\subsection{Do LLMs understand numerical molecular constraints?}

\begin{figure}[!ht]
  \centering
    \centering
    \includegraphics[width=\linewidth]{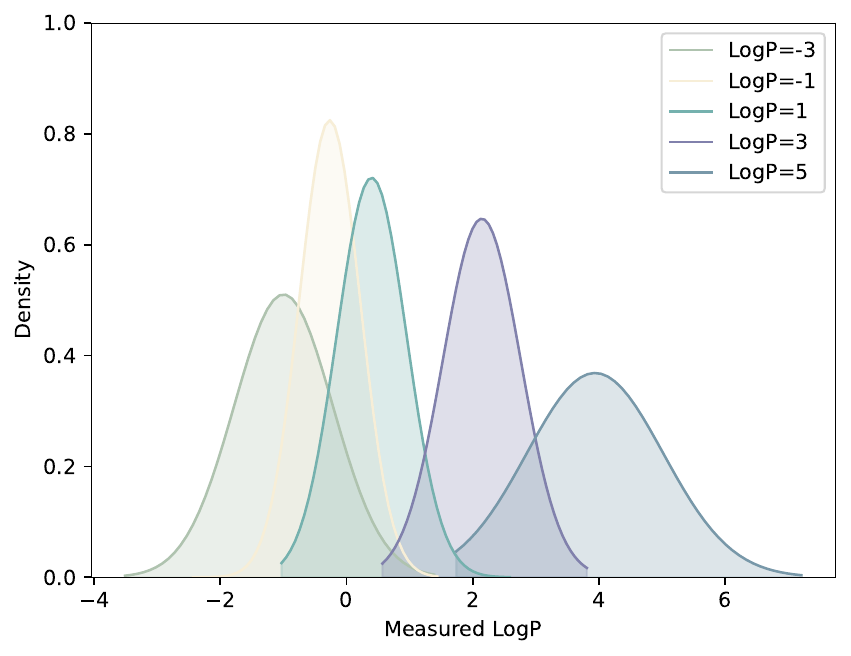}
  \caption{Generated LogP distributions shift systematically toward target values.}
  \label{fig:logp_density}
\end{figure}

\textbf{LLMs can respond to numerical molecular constraints at a distribution level rather than through exact matching.} We use LogP as a case study of numerical molecular constraints. Figure~\ref{fig:logp_density} shows
probability density of generated molecules’ LogP values for different target LogP settings with Gaussian fits. Additional scatter visualizations are provided in Appendix~\ref{app:logp_scatter}. Across all targets, the generated LogP distributions shift consistently with the specified numerical constraints, with the distribution mode moving toward the target value.

This systematic shift indicates that the LLM is incorporating numerical constraints in its generation by adjusting the physicochemical profile of generated molecules in response to the specified value. However, substantial deviations from the target are present, and the generated LogP values exhibit broad dispersion. This suggests that the model does not perform precise numerical control, but rather aligns generation at a coarse, distributional level.

We also compute both Spearman and Pearson correlations and observe strong, statistically significant associations between the target LogP values and the measured LogP of generated molecules (Spearman $\rho = 0.927$, $p < 10^{-6}$; Pearson $r = 0.901$, $p < 10^{-6}$), indicating that the model responds consistently to discrete numerical constraints.

Taken together, these results indicate that numerical molecular constraints are partially understood by LLMs. The models can modulate global property distributions in the intended direction, but they do not reliably generate molecules that closely match a specific numerical target. This behavior highlights a clear capability boundary between qualitative property control and fine-grained numerical accuracy.

\subsection{How do model scale affect creativity?}

\begin{figure}[!ht]
  \centering
  \begin{subfigure}[b]{0.49\textwidth}
    \centering
    \includegraphics[width=\linewidth]{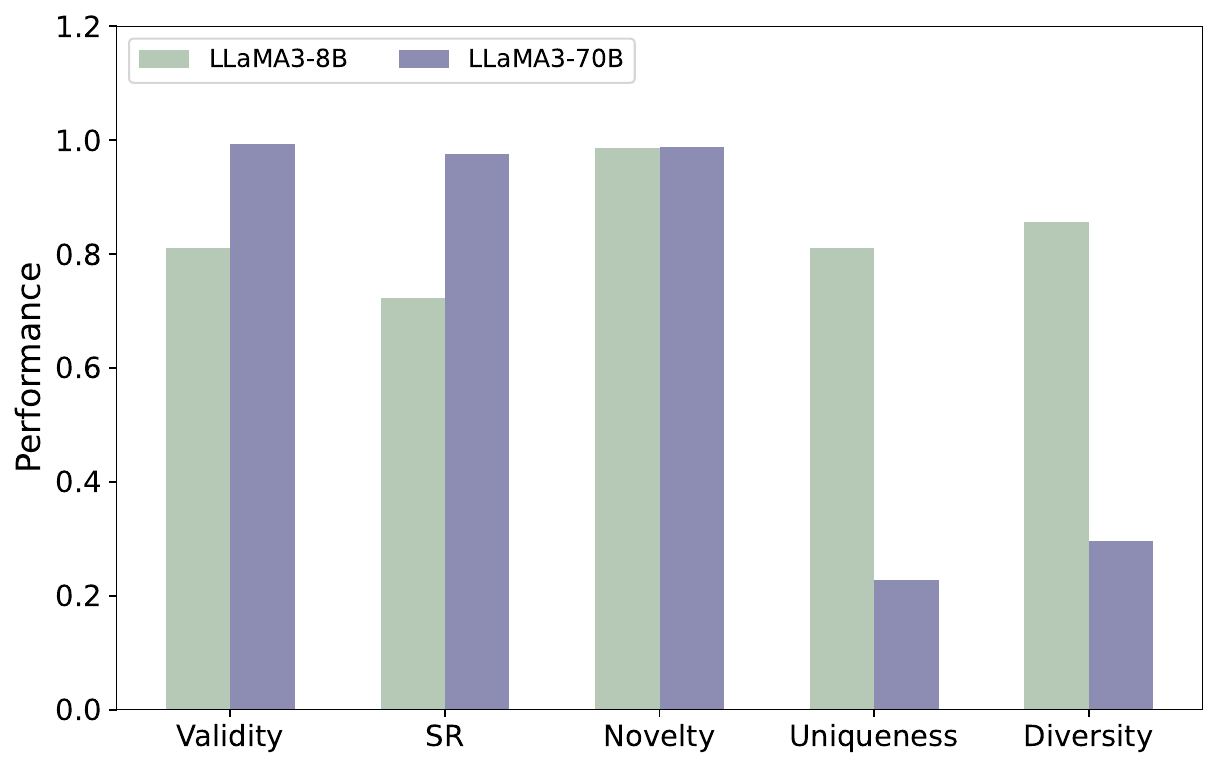}
    \caption{LLaMA3}
    \label{fig:llama3_model_size}
  \end{subfigure}\hfill
  \begin{subfigure}[b]{0.49\textwidth}
    \centering
    \includegraphics[width=\linewidth]{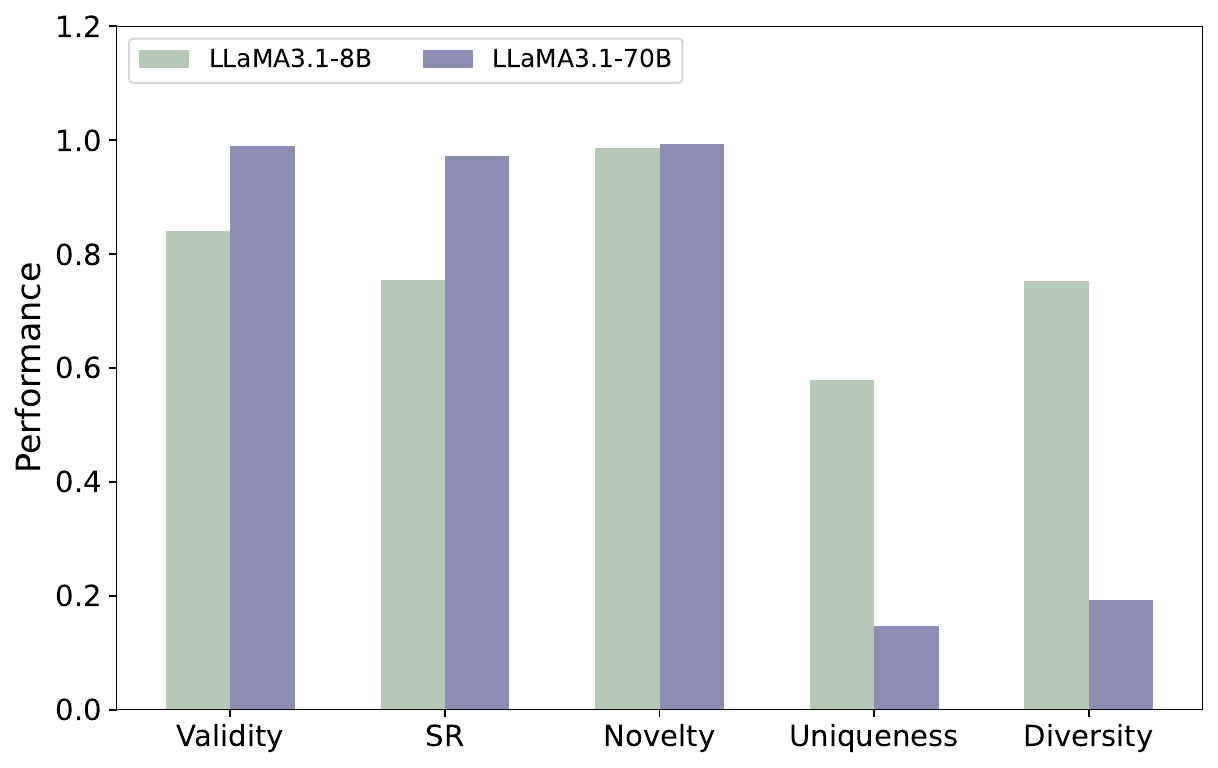}
    \caption{LLaMA3.1}
    \label{fig:llama3.1_model_size}
  \end{subfigure}
  \caption{Larger models achieve higher constraint satisfaction but reduced exploration.}
  \label{fig:model_size}
\end{figure}

\textbf{Increasing model scale improves convergent creativity but reduces divergent creativity.}
Figures~\ref{fig:llama3_model_size} and~\ref{fig:llama3.1_model_size} compare models of different sizes within the LLaMA3 and LLaMA3.1 families. In both cases, larger models show clear improvements in SR and validity. For example, LLaMA3.1-70B consistently outperforms LLaMA3.1-8B, and a similar trend is observed when comparing LLaMA3-70B with LLaMA3-8B. 

At the same time, uniqueness and diversity decrease substantially as model size increases. This indicates that larger models generate molecules that more reliably satisfy constraints, but explore a narrower region of chemical space. Overall, model scaling shifts generation toward stronger convergent creativity at the cost of reduced divergent creativity.

To investigate whether the higher divergent creativity observed in smaller models is due to generating more useful molecules, rather than simply producing more unsuccessful molecules that are therefore diverse, we compare the average Fully Creative across models. This metric evaluates whether the generated molecules satisfy task constraints while remaining novel and unique. The results show that LLaMA3-8B achieves a Fully Creative of 0.556, compared to 0.205 for LLaMA3-70B. A similar trend holds for LLaMA3.1, where the 8B model reaches 0.394, while the 70B model drops to 0.126. These results indicate that the higher divergent creativity of smaller models is not merely driven by failed or low-quality generations, but is associated with a larger number of molecules that are both successful and diverse.

\begin{figure}[!ht]
  \centering
    \centering
    \includegraphics[width=\linewidth]{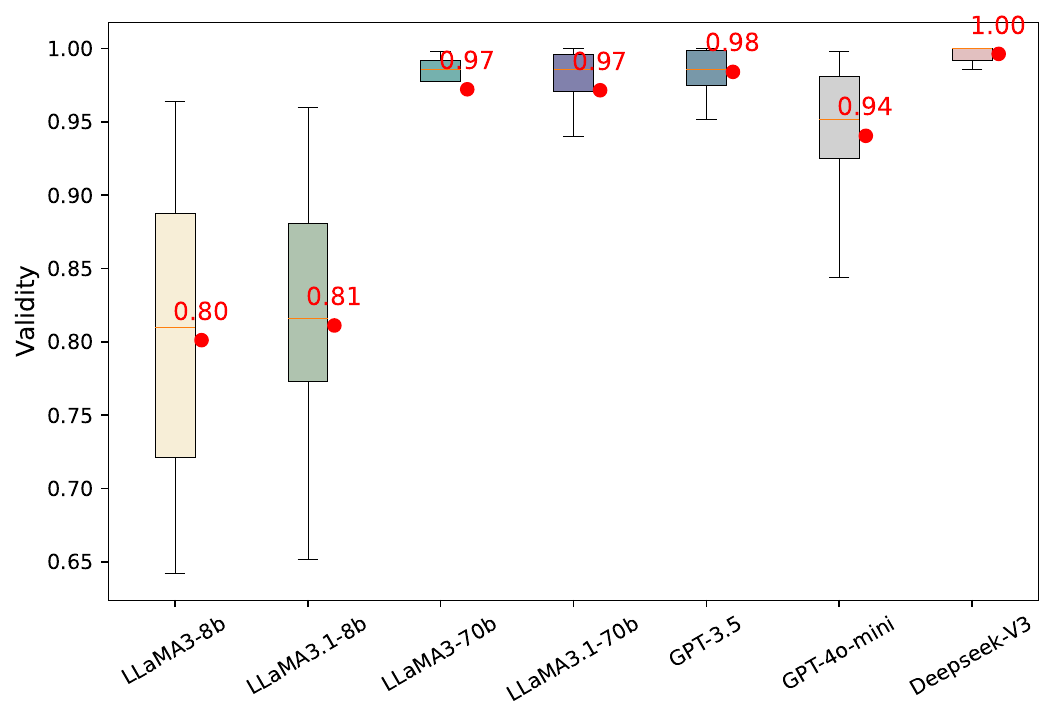}
  \caption{Larger models maintain stable, high validity across diverse constraint settings, while smaller models show substantial variation. The red line represents the median, and the red dot represents the mean.}
      \label{fig:validity_comparison}
\end{figure}

\noindent\textbf{Larger models produce more consistent molecular validity.}
As shown in Figure~\ref{fig:validity_comparison}, we also observe clear differences in the stability of validity across models. Smaller models exhibit substantial variability under different constraint settings. For example, LLaMA3-8B and LLaMA3.1-8B show validity values ranging approximately from 0.64 to 0.96, indicating less stable syntax-level molecular generation. GPT-4o-mini also shows noticeable variation, with validity spanning roughly 0.80 to 0.99.

In contrast, larger models maintain consistently high validity across almost all settings. LLaMA3-70B and LLaMA3.1-70B achieve validity close to 1.00, while GPT-3.5 and DeepSeek-V3 show near-saturated validity with minimal fluctuation. These results indicate that increasing model scale improves the reliability and stability of molecular generation, reinforcing convergent creativity.

\subsection{How do more constraints influence creativity?}

\textbf{Adding compatible constraints can increase success rate rather than reduce it.}
Following \cite{zhou2025instruction, liu2025multi}, we evaluate three two-constraint tasks: QED+SA, BBB+QED, and HIA+QED. Appendix~\ref{app:detailed_results} provides the detailed results. Contrary to the common intuition from natural language generation, where additional constraints usually lower the SR~\cite{li2024control}, we observe that the SRs of these multi-constraint tasks are consistently higher than that of the single QED constraint.

To better understand this behavior, we further compute the QED SR under single SA, BBB, and HIA constraints. As shown in Table~\ref{tab:qed_sr}, most QED SRs under these individual constraints are higher than under the single QED constraint. This indicates that LLMs are particularly effective at satisfying SA, BBB, and HIA constraints, and that molecules generated under these constraints are more likely to satisfy QED as a secondary property.

This phenomenon could be explained by the strong dependencies between chemical property constraints~\cite{pereira2021optimizing}. Unlike domains such as code generation, where atomic operations are largely independent, many molecular properties are correlated through shared structural patterns. Constraints such as BBB permeability or synthetic accessibility bias generation toward specific regions of chemical space that are already enriched for drug-like molecules. As a result, imposing these constraints implicitly increases the probability of satisfying QED, even when QED is not explicitly mentioned in the prompt as a constraint.

Taken together, these results show that adding constraints does not always restrict the generation space in a detrimental way. Instead, when constraints are aligned at the chemical level, they can reshape the generation distribution toward regions that improve convergent creativity, leading to higher success rates despite increased task complexity.

\begin{table}[htbp]
  \centering
    \caption{QED success rate under different constraint settings. The QED SR under single SA, BBB, and HIA constraints is almost always higher than that under the QED constraint.}
    \centering
    \begin{tabular}{cccccccc}
    \toprule
    \textbf{Constraint} & \textbf{Model} & \textbf{QED SR} \\
    \midrule
    \multirow{2}[2]{*}{QED} & LLaMA3-8B & 0.488(0.044) \\
          & LLaMA3.1-8B & 0.498(0.052) \\
    \midrule
    \multirow{2}[2]{*}{SA} & LLaMA3-8B & 0.522(0.047) \\
          & LLaMA3.1-8B & 0.552(0.073) \\
    \midrule
    \multirow{2}[2]{*}{BBB} & LLaMA3-8B & 0.508(0.004) \\
          & LLaMA3.1-8B & 0.616(0.024) \\
    \midrule
    \multirow{2}[2]{*}{HIA} & LLaMA3-8B & 0.38(0.073) \\
          & LLaMA3.1-8B & 0.542(0.065) \\
    \bottomrule
    \end{tabular}%
  \label{tab:qed_sr}%
\end{table}%

\begin{table*}[!tbp]
\caption{Effect of in-context learning on biological activity tasks. In-context learning substantially improves success rate but reduces exploration, yielding only marginal gains in Fully Creative. ZS denotes zero-shot, and ICL denotes in-context learning.}
\centering
\begin{tabular}{l|c|c|c|c|c|c}
\toprule
\textbf{Task} & \textbf{Validity} & \textbf{SR} & \textbf{Novelty} & \textbf{Uniqueness} & \textbf{Diversity} & \textbf{Fully Creative} \\
\midrule
DRD2 \std{ZS} & 0.978 \std{0.015} & 0.15 \std{0.019} & 0.981 \std{0.014} & 0.957 \std{0.018} & 0.879 \std{0.006} & 0.15 \std{0.019} \\
\rowcolor{tablegreen}
DRD2 \std{ICL} & 0.9 \std{0.056} & 0.542 \std{0.047} & 0.876 \std{0.025} & 0.688 \std{0.038} & 0.83 \std{0.008} & 0.19 \std{0.021} \\
\midrule
JNK3 \std{ZS} & 0.976 \std{0.013} & 0.006 \std{0.005} & 0.972 \std{0.012} & 0.951 \std{0.009} & 0.86 \std{0.005} & 0.006 \std{0.005} \\
\rowcolor{tablegreen}
JNK3 \std{ICL} & 0.732 \std{0.019} & 0.316 \std{0.048} & 0.902 \std{0.026} & 0.763 \std{0.069} & 0.848 \std{0.015} & 0.092 \std{0.028} \\
\midrule
GSK3$\beta$ \std{ZS} & 0.952 \std{0.016} & 0.032 \std{0.019} & 0.976 \std{0.024} & 0.965 \std{0.012} & 0.875 \std{0.004} & 0.032 \std{0.019} \\
\rowcolor{tablegreen}
GSK3$\beta$ \std{ICL} & 0.84 \std{0.041} & 0.698 \std{0.028} & 0.737 \std{0.045} & 0.406 \std{0.027} & 0.643 \std{0.039} & 0.112 \std{0.033} \\
\bottomrule
\end{tabular}
\label{tab:icl_results}
\end{table*}

\subsection{Does in-context learning improve creativity?}

\textbf{In-context learning boosts convergent creativity but reduces divergent creativity.}
We evaluate the effect of in-context learning (ICL) on biological target–specific generation using GPT-3.5 for three targets: DRD2, JNK3, and GSK3$\beta$. For each target, we compare zero-shot (ZS) generation with a 10-shot ICL setting.

ICL examples are selected as follows. We first identify molecules whose activity values fall within the top 10\% of all active molecules. These molecules are then clustered based on Tanimoto similarity, and the medoid of each cluster is selected. The medoids from ten clusters are used as the 10-shot ICL examples, ensuring both high activity and structural coverage.

As shown in Table~\ref{tab:icl_results}, across all three targets, a consistent pattern emerges. In the zero-shot setting, the SR remains very low—0.150 for DRD2, 0.006 for JNK3, and 0.032 for GSK3$\beta$—despite high validity (above 0.95) and high diversity (above 0.86). This indicates that, without explicit examples, the model can generate chemically valid and structurally diverse molecules, but fails to satisfy target-specific bioactivity constraints. 

Introducing 10-shot ICL leads to a substantial increase in bioactivity satisfaction. The SR rises to 0.542 for DRD2, 0.316 for JNK3, and 0.698 for GSK3$\beta$. This shows that a small number of active examples enables the model to internalize task-specific structure–activity relationships and guides generation toward a more relevant chemical region.

However, this improvement in SR is accompanied by clear trade-offs. Validity decreases under ICL, most notably for JNK3, where it drops from 0.976 to 0.732. More importantly, divergent creativity metrics decline sharply. For GSK3$\beta$, uniqueness drops from 0.965 to 0.406, and novelty decreases from 0.976 to 0.737. These changes indicate that ICL strongly narrows the explored chemical space.

This effect is further reflected by the Fully Creative metric, which measures the proportion of molecules that are simultaneously successful, unique, and novel. Although SR increases substantially under ICL—for example, from 0.150 to 0.542 for DRD2—the corresponding Fully Creative value increases only marginally, from 0.150 to 0.190. This mismatch shows that most additional successful molecules introduced by ICL are structurally redundant.

Overall, these results indicate that ICL improves convergent creativity by increasing the likelihood of satisfying biological constraints, but does so by reducing divergent creativity. The model relies heavily on structural patterns present in the provided examples, leading to repetitive generations and limited exploration beyond the example-defined region.

\subsection{How does sampling temperature affect the trade-off?}

\textbf{Increasing the temperature can enhance exploration, while excessive temperature leads to degraded constraint satisfaction.}
We study the effect of sampling temperature on molecular generation using LLaMA3-8B under the SA constraint. The temperature is varied from 0.25 to 1.5, covering a spectrum from conservative sampling to highly stochastic generation.

\begin{figure}[!ht]
  \centering
    \centering
    \includegraphics[width=\linewidth]{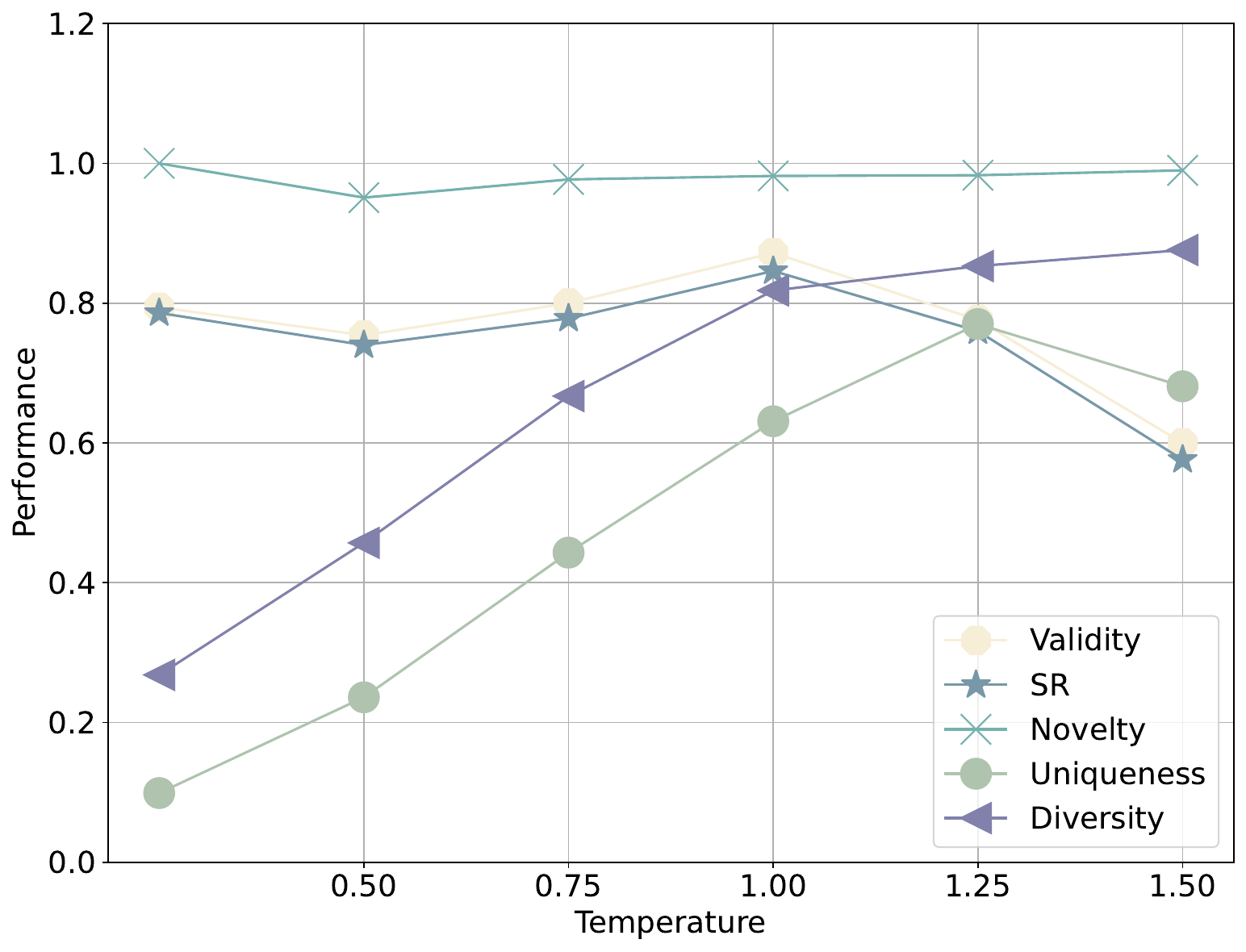}
  \caption{Sampling temperature enables inference-time control of the creativity trade-off. Moderate increases (0.25→1.0) improve diversity without sacrificing validity, but excessive temperature degrades constraint satisfaction.}
  \label{fig:temperature}
\end{figure}

As shown in Figure~\ref{fig:temperature}, divergent creativity increases consistently with temperature. Both uniqueness and diversity rise as temperature increases, indicating broader exploration of chemical space. Diversity shows a near-monotonic growth over the entire range, while uniqueness increases rapidly up to temperature 1.25 and then slightly decreases at 1.5, suggesting diminishing returns in structural variety at very high temperature.

In contrast, convergent creativity remains relatively stable only within a moderate temperature range. From 0.25 to 1.0, SR and validity do not deteriorate and even show mild improvement, reaching their peak around temperature 1.0. However, when the temperature is further increased to 1.25 and 1.5, both SR and validity decline noticeably, indicating reduced constraint satisfaction under highly stochastic sampling.

Overall, these results show that sampling temperature provides an effective inference-time mechanism for controlling the balance between convergent and divergent creativity. Moderate temperature increase enhances exploration without degrading success, but excessively high temperature leads to a clear drop in constraint satisfaction. This observation suggests the existence of a practical temperature range in which diversity can be improved without sacrificing generation reliability, without modifying the model or retraining.

\subsection{How to find the most creative LLM?}

Our framework defines \textbf{Overall Creativity} as a unified metric that combines convergent and divergent creativity, enabling a direct comparison of how creative different LLMs are at the system level. Appendix~\ref{app:overall_creativity} compares the average Overall Creativity of the analyzed LLMs across physicochemical, ADMET, activity, and numerical constraint tasks. GPT-3.5 achieves the highest Overall Creativity on physicochemical, ADMET, and activity tasks, while DeepSeek-V3 performs best on the numerical constraint task.

\section{Implications for Molecular Design}

Our results suggest practical guidelines for using LLMs in molecular generation, clarifying when they are effective and how to balance reliability with exploration. More detailed guidelines are provided in Appendix~\ref{app:guidelines}.

\noindent \textbf{Task alignment matters.} LLMs perform reliably on tasks driven by global molecular properties (drug-likeness, synthetic accessibility, ADMET), which align with patterns learned during pretraining. However, for target-specific activity tasks, high validity and diversity can coexist with near-zero success rates, creating misleading progress signals. LLM outputs in such settings should not be treated as reliable inhibitor candidates without validation.

\noindent \textbf{Model size and constraint design.} Larger models consistently achieve higher validity and success rates, making them preferable when reliability is prioritized. Applying multiple cooperative constraints improves molecular quality by guiding generation toward chemically meaningful regions. LLMs can also respond to explicit numerical constraints for numerical property control (e.g., LogP value).

\noindent \textbf{Sampling and in-context learning.} Moderately increasing temperature improves diversity while preserving constraint satisfaction, though excessive temperatures degrade success rates. In-context learning primarily boosts constraint satisfaction but has limited impact on novelty and diversity, sometimes encouraging replication of provided examples rather than exploration.

\section{Conclusion}

In this work, we systematically examined the creative behavior of large language models in molecule generation. By framing creativity as a functional requirement rather than an aesthetic concept, we introduced a two-dimensional evaluation framework that distinguishes between convergent creativity, which reflects constraint satisfaction, and divergent creativity, which reflects exploration. Through extensive experiments across physicochemical, ADMET, and biological activity tasks, we showed that LLMs exhibit distinct and task-dependent creative behaviors in molecular generation. Our findings clarify both the strengths and limitations of current LLMs and suggest the appropriate use of LLMs in molecular discovery pipelines.

\section*{Impact Statement}

This paper presents work whose goal is to advance the field of machine learning. There are many potential societal consequences of our work, none of which we feel must be specifically highlighted here.


\bibliography{icml2026}
\bibliographystyle{icml2026}

\newpage
\appendix
\onecolumn


\section{Related Work}
\label{app:related_work}

\paragraph{Molecule Generation.} 
Deep generative models for molecule generation include variational autoencoders (VAEs) \cite{gomez2018automatic, simonovsky2018graphvae}, generative adversarial networks (GANs) \cite{guimaraes2017objective, de2018molgan}, normalizing flow models \cite{shi2020graphaf, song2023equivariant}, autoregressive models \cite{popova2019molecularrnn, gebauer2019symmetry}, and diffusion-based models \cite{hoogeboom2022equivariant, xu2022geodiff}. These approaches differ in molecular representations and generation mechanisms. To evaluate these generative models, a set of standard metrics has been widely adopted \cite{polykovskiy2020molecular}. In existing work, these metrics are primarily used as performance indicators to compare models or training strategies. Their role is largely descriptive, where higher values are interpreted as better generation quality. In contrast, our work reorganizes standard evaluation metrics into a creativity-oriented framework by treating constraint satisfaction and exploration as different aspects of generation behavior, and by analyzing how convergent and divergent creativity co-exist and compete when LLMs are used as molecular generators, beyond conventional performance comparisons.

\paragraph{LLM Creativity Evaluation.}
Creativity evaluation of large language models has been studied in open-ended generation tasks such as story writing \cite{atmakuru2024cs4, chakrabarty2024art}, abstract generation \cite{lu2024ai}, role-play and dialogue scenarios \citep{lu2024llm}, as well as problem-solving tasks including physical reasoning \cite{tian2024macgyver} and code generation \cite{delorenzo2024creativeval, lu2025benchmarking}. Creativity in these works is often quantified using subcomponents derived from the Torrance Tests of Creative Thinking \cite{torrance1966torrance}, including fluency, flexibility, originality, and elaboration. Depending on the task, these dimensions are measured through lexical diversity, semantic distance, solution variety, or human preference judgments. Despite this progress, creativity evaluation in molecule generation remains underexplored. Molecules are subject to strict chemical validity and functional constraints, which fundamentally differentiate this domain from text or code generation. As a result, creativity cannot be assessed through methods used in existing work, and a systematic analysis is still missing. Our work fills this gap by treating creativity as an emergent property arising from the interaction between exploration and constraint satisfaction, rather than as a single scalar score. To our knowledge, this is the first work that systematically evaluates creativity in LLM-based molecule generation from this perspective.

\section{Limitations}

Despite the systematic evaluation presented in this work, several limitations remain. Being explicit about these issues is important, both for correctly interpreting the results and for guiding future research on LLM-based molecular generation.

\textbf{Ambiguity of creativity metrics.}
A first limitation lies in how creativity is operationalized. Although we decompose creativity into convergent and divergent components using measurable indicators, these metrics are still proxies rather than direct measurements of chemical creativity. Different choices of metrics can lead to different quantitative conclusions. As a result, current creativity metrics capture only partial aspects of what chemists would intuitively regard as creative molecular design.

\textbf{Dependence on the training corpus.}
LLM behavior is strongly shaped by the chemical patterns present in their training data. Many global properties, such as drug-likeness or simple physicochemical trends, can be inferred from frequent motifs and correlations in large text or SMILES corpora. In contrast, protein-specific activity often depends on subtle structure–activity relationships that are sparsely represented or inconsistently annotated in public data.

\section{Detailed Guidelines}
\label{app:guidelines}

Based on our experimental results, we summarize the following practical guidelines for using LLMs in molecular generation. These guidelines clarify when LLMs are effective, where their limitations lie, and how to configure them to balance accuracy and exploration.

\noindent \textbf{Choose task types aligned with global molecular constraints.}  
LLMs perform reliably in molecular generation tasks driven by coarse-grained and global molecular properties, such as drug-likeness, synthetic accessibility, and ADMET-related constraints. These tasks are well aligned with the statistical patterns learned during pretraining, enabling LLMs to generate valid and constraint-compliant molecules efficiently.

\noindent \textbf{Avoid treating LLM outputs as reliable target-specific inhibitors.}  
For target-specific activity optimization, high validity and diversity can coexist with near-zero success rate. This discrepancy can create misleading signals of progress. Even when in-context learning improves success rates, the accompanying loss in novelty and uniqueness indicates a tendency to overfit to provided examples. LLM-generated molecules in such settings should therefore not be interpreted as reliable candidates without further validation.

\noindent \textbf{Use LLMs as proposal generators rather than decision makers.}  
LLMs are best used to propose chemically plausible candidates, while downstream evaluation should rely on external modules such as property predictors, docking or scoring functions, and physics-based filters. This separation reduces the risk of redundant candidates and poor generalization.

\noindent \textbf{Prefer larger LLMs when higher constraint satisfaction is required.}  
Our results show that larger models consistently achieve higher validity and success rates under constrained generation settings, making them more suitable when reliability is prioritized over exploration.

\noindent \textbf{Combine compatible constraints to improve molecular quality.}  
Applying multiple cooperative constraints can guide generation toward chemically meaningful regions of chemical space, improving overall molecular quality without relying on overly strict single constraints.

\noindent \textbf{Adjust sampling temperature to balance diversity and accuracy.}  
Moderately increasing the sampling temperature improves diversity while largely preserving constraint satisfaction. However, excessively high temperatures tend to degrade success rate and should be avoided.

\noindent \textbf{Use numerical constraints when numerical property control is required.}  
LLMs can respond to explicit numerical molecular constraints, making them capable of handling tasks that require controlling properties such as LogP within specified ranges.

\noindent \textbf{Interpret in-context learning primarily as an accuracy booster.}  
In-context learning mainly improves constraint satisfaction and success rate. Its impact on novelty and diversity is limited, and in some cases it encourages replication of provided examples.

\section{Evaluation Metrics}
\label{app:metrics}

We provide formal definitions for all metrics used to evaluate LLM creativity in molecular generation. Let $G$ denote the total number of generated molecules and $R$ denote the reference dataset.

\subsection{Convergent Creativity Metrics}

\noindent\textbf{Validity.} Measures whether generated SMILES strings correspond to chemically valid molecular structures. We use RDKit~\cite{landrum2013rdkit} to verify the validity. Let $G$ denote the total number of generated molecules and $V$ denote the number of molecules successfully parsed by RDKit:
\begin{equation}
\text{Validity} = \frac{V}{G}.
\end{equation}

\noindent\textbf{Success Rate (SR).} Measures the proportion of generated molecules that satisfy all task-specific constraints. Let $S$ denote the number of valid molecules that meet all specified constraints. For multi-constraint tasks, a molecule is considered successful only if it satisfies \emph{all} constraints simultaneously:
\begin{equation}
\text{Success Rate} = \frac{S}{G}.
\end{equation}

\subsection{Divergent Creativity Metrics}

\noindent\textbf{Novelty.} Measures the fraction of generated molecules that do not appear in the reference dataset $R$. Following prior work~\citep{gao2022sample, li2024tomg}, we use ZINC 250K~\citep{sterling2015zinc} as the reference dataset. Let $N$ denote the number of valid molecules not present in $R$:
\begin{equation}
\text{Novelty} = \frac{N}{V}.
\end{equation}

\noindent\textbf{Uniqueness.} Measures the proportion of non-duplicate molecules among valid generated molecules. This metric ensures the model does not collapse to producing only a few repetitive structures. Let $U$ denote the number of unique molecules (after removing duplicates):
\begin{equation}
\text{Uniqueness} = \frac{U}{V}.
\end{equation}

\noindent\textbf{Diversity.} Quantifies the average structural dissimilarity among generated valid molecules. We compute pairwise Tanimoto similarity using Morgan fingerprints~\cite{rogers2010extended} (radius 3, 2048 bits), then aggregate across all pairs. Let $V = \{m_1, \ldots, m_n\}$ denote the set of valid molecules, and $\text{sim}(m_i, m_j)$ denote the Tanimoto similarity between molecules $m_i$ and $m_j$:
\begin{equation}
\text{Diversity} = 1 - \frac{2}{n(n-1)} \sum_{1 \le i < j \le n} \text{sim}(m_i, m_j).
\end{equation}
Higher diversity indicates broader exploration of chemical space.

\subsection{Composite Metrics}

\noindent\textbf{Convergent Creativity.} Aggregates validity and success rate using their geometric mean:
\begin{equation}
\text{Convergent Creativity} = \sqrt{\text{Validity} \times \text{Success Rate}}.
\end{equation}

\noindent\textbf{Divergent Creativity.} Aggregates novelty, uniqueness, and diversity using their geometric mean:
\begin{equation}
\text{Divergent Creativity} = \left(\text{Novelty} \times \text{Uniqueness} \times \text{Diversity}\right)^{1/3}.
\end{equation}

\noindent\textbf{Overall Creativity.} Combines convergent and divergent creativity:
\begin{equation}
\text{Overall Creativity} = \sqrt{\text{Convergent Creativity} \times \text{Divergent Creativity}}.
\end{equation}
The geometric mean ensures that both dimensions must be reasonably high to achieve a strong overall score, preventing models from excelling on only one dimension.

\subsection{Elite Subset Metric}

\noindent\textbf{Fully Creative.} Measures the proportion of molecules that are simultaneously successful (satisfy all task constraints), novel (not in reference dataset), and unique (non-duplicate within the batch). Let $E$ denote the number of molecules satisfying all three conditions:
\begin{equation}
\text{Fully Creative} = \frac{E}{G}.
\end{equation}
This metric identifies molecules that represent genuine creative discoveries, filtering out redundant or previously known structures.

\subsection{Theoretical Justification for Geometric Mean Aggregation}

We justify our choice of geometric mean for aggregating metrics through three complementary perspectives: mathematical properties, detection of degenerate solutions, and balanced optimization incentives.

\subsubsection{Why Geometric Mean Over Arithmetic Mean?}

The geometric mean possesses desirable properties for evaluating creativity that the arithmetic mean lacks. For metrics $x_1, \ldots, x_n \in [0,1]$, the geometric mean is defined as:
\begin{equation}
\text{GM}(x_1, \ldots, x_n) = \left(\prod_{i=1}^n x_i\right)^{1/n}.
\end{equation}

\noindent\textbf{Property 1: Penalizes imbalanced performance.} The geometric mean is strictly dominated by the arithmetic mean:
\begin{equation}
\text{GM}(x_1, \ldots, x_n) \le \text{AM}(x_1, \ldots, x_n) = \frac{1}{n}\sum_{i=1}^n x_i,
\end{equation}
with equality if and only if $x_1 = \cdots = x_n$. This means models cannot achieve high scores by excelling on only one dimension while neglecting others.

\noindent\textbf{Property 2: Enforces zero-failure principle.} If any component metric is zero, the geometric mean becomes zero:
\begin{equation}
x_i = 0 \text{ for any } i \implies \text{GM}(x_1, \ldots, x_n) = 0.
\end{equation}
This is critical for molecular generation: a model generating entirely invalid molecules (Validity=0) or completely redundant outputs (Uniqueness=0) should receive zero credit, regardless of other metrics.

\noindent\textbf{Property 3: Scale-invariant sensitivity.} The geometric mean treats multiplicative changes uniformly. For a model with current scores $(x_1, \ldots, x_n)$, improving any metric $x_i$ by a factor $\alpha$ yields the same relative improvement in GM:
\begin{equation}
\frac{\text{GM}(x_1, \ldots, \alpha x_i, \ldots, x_n)}{\text{GM}(x_1, \ldots, x_n)} = \alpha^{1/n}.
\end{equation}
This prevents disproportionate focus on already high metrics (as would occur with arithmetic mean).

\subsubsection{Convergent Creativity: Why Validity and Success Rate?}

Convergent creativity measures constraint satisfaction through two distinct but complementary aspects:

\noindent\textbf{Validity} captures \emph{syntactic correctness}—whether the model understands basic chemical grammar (valence rules, aromaticity, bond consistency). A model generating invalid SMILES has failed at the most fundamental level.

\noindent\textbf{Success Rate} captures \emph{semantic correctness}—whether the model satisfies task-specific constraints (e.g., QED $\ge 0.6$, DRD2 binding $\ge 0.5$). High validity alone is insufficient; molecules must also be \emph{functionally relevant}.

The geometric mean enforces that both must be high:
\begin{equation}
\text{Convergent Creativity} = \sqrt{\text{Validity} \times \text{Success Rate}}.
\end{equation}

Consider one degenerate case:
\begin{itemize}
\item \textbf{Case A:} Validity=1.0, SR=0.0 $\implies$ Convergent Creativity=0.0. The model produces chemically valid molecules but none satisfy the task constraints—complete task failure.
\end{itemize}

Using arithmetic mean would give Case A a score of 0.5, masking these critical failures. The geometric mean correctly identifies this case as inadequate.

\subsubsection{Divergent Creativity: Why Novelty, Uniqueness, and Diversity?}

Divergent creativity measures exploration through three non-redundant dimensions:

\noindent\textbf{Novelty} measures exploration beyond the \emph{training distribution}. Without novelty, the model merely memorizes known molecules rather than discovering new ones.

\noindent\textbf{Uniqueness} measures exploration within the \emph{generated batch}. Even if all molecules are novel, generating 100 near-identical structures indicates mode collapse, not creative exploration.

\noindent\textbf{Diversity} measures \emph{structural dissimilarity}. Two molecules can be unique (different SMILES) yet structurally similar (e.g., differing by one methyl group). Diversity captures broader exploration of chemical scaffolds.

These three metrics capture distinct failure modes:
\begin{equation}
\text{Divergent Creativity} = (\text{Novelty} \times \text{Uniqueness} \times \text{Diversity})^{1/3}.
\end{equation}

Consider degenerate cases:
\begin{itemize}
\item \textbf{Memorization:} Novelty=0, Uniqueness=1.0, Diversity=0.8 $\implies$ Divergent Creativity=0. The model reproduces training data—no discovery occurs.
\item \textbf{Mode collapse:} Novelty=1.0, Uniqueness=0.1, Diversity=0.1 $\implies$ Divergent Creativity=$\approx 0.22$. Despite generating novel molecules, the model explores only a tiny region.
\item \textbf{Insufficient scaffold exploration:} Novelty=1.0, Uniqueness=1.0, Diversity=0.3 $\implies$ Divergent Creativity=$\approx 0.67$. All molecules are novel and unique, but structurally homogeneous.
\end{itemize}

The geometric mean ensures all three dimensions must be reasonably high, preventing models from achieving high scores through narrow exploration strategies.

\subsubsection{Overall Creativity: Balancing Convergent and Divergent}

Molecular generation requires \emph{both} constraint satisfaction and exploration. A model that generates valid, task-compliant molecules but only reproduces training data has failed. Conversely, a model that explores broadly but produces mostly unusable molecules has also failed.

We define Overall Creativity as:
\begin{equation}
\text{Overall Creativity} = \sqrt{\text{Convergent Creativity} \times \text{Divergent Creativity}}.
\end{equation}

This formulation enforces that models must balance both objectives:

\noindent\textbf{Why not maximize them separately?} Treating convergent and divergent creativity as independent objectives would require multi-objective optimization (e.g., Pareto frontiers). The geometric mean provides a single scalar that encourages balanced improvement.

\noindent\textbf{Detecting unbalanced models.} Consider three hypothetical models:
\begin{align}
\text{Model A:}\quad &\text{Conv}=0.9, \text{Div}=0.1 \implies \text{Overall}=0.3 \\
\text{Model B:}\quad &\text{Conv}=0.5, \text{Div}=0.5 \implies \text{Overall}=0.5 \\
\text{Model C:}\quad &\text{Conv}=0.7, \text{Div}=0.7 \implies \text{Overall}=0.7
\end{align}

Model A excels at constraint satisfaction but barely explores—likely overfitting to narrow chemical patterns. Model B is mediocre at both. Model C achieves balanced competence. The geometric mean correctly ranks $C > B > A$, while arithmetic mean would rank $C > A = B$, failing to detect the imbalanced model.

\noindent\textbf{Formal optimality condition.} For fixed product $p = \text{Conv} \times \text{Div}$, the geometric mean $\sqrt{p}$ is maximized when $\text{Conv} = \text{Div}$. This can be shown via Lagrange multipliers or AM-GM inequality. Thus, Overall Creativity naturally incentivizes balanced development of both creativity dimensions.

\subsubsection{Empirical Validation of Non-Redundancy}

Our correlation analysis (Table~\ref{tab:pearson_corr_metrics}) provides empirical evidence that these aggregations capture distinct information:

\begin{itemize}
\item \textbf{Within convergent:} Validity and SR are highly correlated ($r=0.99$), but not redundant—they capture syntax vs. semantics.
\item \textbf{Within divergent:} Novelty, Uniqueness, and Diversity are correlated ($r=0.86$--$0.96$), but this reflects a shared exploratory tendency rather than measurement redundancy. Each can vary independently in edge cases (as shown in the degenerate examples above).
\item \textbf{Between dimensions:} Convergent and divergent metrics are strongly anti-correlated ($r=-0.70$ to $-0.89$), confirming they measure competing objectives that must be balanced.
\end{itemize}

This anti-correlation validates the need for an aggregation scheme that prevents one dimension from dominating the other—precisely what geometric mean achieves.

\subsection{Implementation Details}

All metrics are computed over five independent runs, each generating 100 molecules. We report mean values with standard deviations in parentheses. Validity and Success Rate are computed over all generated molecules, while all other metrics are computed only over valid molecules. For Tanimoto similarity calculations, we use RDKit's Morgan fingerprint implementation with default parameters (radius=3, nBits=2048). Reference dataset ZINC 250K~\cite{sterling2015zinc} is preprocessed to canonical SMILES representation to ensure consistent comparison. We also include the DRD2~\cite{nakamura2025molecular}, JNK3, and GSK3$\beta$~\cite{jin2020multi} datasets in the reference dataset, as they are used for selecting in-context learning samples.

\section{Detailed Results}

\label{app:detailed_results}

In this section, we present the detailed results of all models across all tasks in Tables~\ref{tab:results_llama3-8b} --~\ref{tab:results_deepseek-chat-v3-0324}. The reported values are the mean and standard deviation over five independent runs. CC denotes Convergent Creativity, and DC denotes Divergent Creativity. For the LogP task, we consider a generated molecule to be successful if its LogP differs from the specified value by at most 1.

\begin{table}[!b]
\caption{Detailed results for LLaMA3-8B}
\label{tab:results_llama3-8b}
\resizebox{\textwidth}{!}{
\begin{tabular}{lccccccccc}
\toprule
Task & Validity & SR & Novelty & Uniqueness & Diversity & CC & DC & Creativity & Fully Creative \\
\midrule
BBB & 0.82 \std{0.051} & 0.774 \std{0.043} & 0.989 \std{0.006} & 0.882 \std{0.033} & 0.883 \std{0.014} & 0.797 \std{0.046} & 0.917 \std{0.014} & 0.854 \std{0.027} & 0.67 \std{0.04} \\
HIA & 0.828 \std{0.051} & 0.78 \std{0.05} & 0.981 \std{0.006} & 0.776 \std{0.039} & 0.834 \std{0.029} & 0.804 \std{0.047} & 0.859 \std{0.024} & 0.831 \std{0.033} & 0.586 \std{0.055} \\
\midrule
QED & 0.724 \std{0.074} & 0.488 \std{0.044} & 0.99 \std{0.023} & 0.953 \std{0.007} & 0.888 \std{0.008} & 0.594 \std{0.057} & 0.943 \std{0.01} & 0.748 \std{0.04} & 0.452 \std{0.055} \\
SA & 0.872 \std{0.029} & 0.846 \std{0.032} & 0.982 \std{0.015} & 0.631 \std{0.066} & 0.818 \std{0.031} & 0.859 \std{0.03} & 0.797 \std{0.038} & 0.827 \std{0.033} & 0.516 \std{0.075} \\
\midrule
DRD2 & 0.688 \std{0.041} & 0.018 \std{0.011} & 0.997 \std{0.007} & 0.997 \std{0.006} & 0.907 \std{0.004} & 0.107 \std{0.034} & 0.966 \std{0.003} & 0.319 \std{0.05} & 0.018 \std{0.011} \\
JNK3 & 0.642 \std{0.033} & 0.0 \std{0.0} & 0.994 \std{0.008} & 1.0 \std{0.0} & 0.898 \std{0.006} & 0.0 \std{0.0} & 0.963 \std{0.003} & 0.0 \std{0.0} & 0.0 \std{0.0} \\
GSK3 & 0.642 \std{0.082} & 0.002 \std{0.004} & 1.0 \std{0.0} & 0.994 \std{0.009} & 0.903 \std{0.003} & 0.018 \std{0.039} & 0.965 \std{0.003} & 0.058 \std{0.13} & 0.002 \std{0.004} \\
\midrule
BBB,QED & 0.766 \std{0.038} & 0.508 \std{0.058} & 0.997 \std{0.006} & 0.99 \std{0.016} & 0.902 \std{0.007} & 0.623 \std{0.049} & 0.962 \std{0.006} & 0.774 \std{0.028} & 0.504 \std{0.054} \\
HIA,QED & 0.718 \std{0.045} & 0.492 \std{0.042} & 0.985 \std{0.011} & 0.967 \std{0.023} & 0.9 \std{0.008} & 0.594 \std{0.041} & 0.95 \std{0.009} & 0.751 \std{0.028} & 0.462 \std{0.049} \\
QED,SA & 0.81 \std{0.032} & 0.568 \std{0.05} & 0.978 \std{0.007} & 0.783 \std{0.082} & 0.828 \std{0.036} & 0.678 \std{0.04} & 0.858 \std{0.042} & 0.762 \std{0.028} & 0.384 \std{0.063} \\
\midrule
LogP,-1 & 0.94 \std{0.016} & 0.416 \std{0.03} & 0.995 \std{0.01} & 0.468 \std{0.064} & 0.774 \std{0.025} & 0.625 \std{0.024} & 0.711 \std{0.039} & 0.666 \std{0.024} & 0.176 \std{0.048} \\
LogP,-3 & 0.924 \std{0.032} & 0.01 \std{0.007} & 0.991 \std{0.014} & 0.629 \std{0.087} & 0.823 \std{0.024} & 0.084 \std{0.05} & 0.799 \std{0.044} & 0.231 \std{0.131} & 0.008 \std{0.004} \\
LogP,1 & 0.964 \std{0.019} & 0.758 \std{0.013} & 0.98 \std{0.014} & 0.542 \std{0.051} & 0.837 \std{0.01} & 0.855 \std{0.008} & 0.763 \std{0.029} & 0.807 \std{0.016} & 0.34 \std{0.042} \\
LogP,3 & 0.904 \std{0.022} & 0.222 \std{0.048} & 0.986 \std{0.001} & 0.81 \std{0.043} & 0.855 \std{0.016} & 0.446 \std{0.051} & 0.881 \std{0.021} & 0.626 \std{0.036} & 0.212 \std{0.043} \\
LogP,5 & 0.776 \std{0.046} & 0.106 \std{0.036} & 0.983 \std{0.007} & 0.918 \std{0.02} & 0.875 \std{0.01} & 0.282 \std{0.052} & 0.924 \std{0.009} & 0.509 \std{0.047} & 0.106 \std{0.036} \\
\bottomrule
\end{tabular}
}
\end{table}

\begin{table}
\caption{Detailed results for LLaMA3.1-8B}
\label{tab:results_llama3.1-8b}
\resizebox{\textwidth}{!}{
\begin{tabular}{lccccccccc}
\toprule
Task & Validity & SR & Novelty & Uniqueness & Diversity & CC & DC & Creativity & Fully Creative \\
\midrule
BBB & 0.888 \std{0.026} & 0.87 \std{0.019} & 0.987 \std{0.019} & 0.538 \std{0.046} & 0.74 \std{0.03} & 0.879 \std{0.02} & 0.732 \std{0.03} & 0.802 \std{0.019} & 0.454 \std{0.031} \\
HIA & 0.816 \std{0.051} & 0.794 \std{0.045} & 0.984 \std{0.018} & 0.59 \std{0.059} & 0.733 \std{0.032} & 0.805 \std{0.048} & 0.752 \std{0.035} & 0.777 \std{0.032} & 0.452 \std{0.058} \\
\midrule
QED & 0.78 \std{0.037} & 0.498 \std{0.052} & 0.977 \std{0.015} & 0.805 \std{0.069} & 0.849 \std{0.031} & 0.623 \std{0.041} & 0.874 \std{0.037} & 0.737 \std{0.021} & 0.354 \std{0.049} \\
SA & 0.874 \std{0.037} & 0.858 \std{0.038} & 0.995 \std{0.012} & 0.383 \std{0.056} & 0.69 \std{0.062} & 0.866 \std{0.037} & 0.639 \std{0.044} & 0.743 \std{0.023} & 0.316 \std{0.043} \\
\midrule
DRD2 & 0.652 \std{0.043} & 0.014 \std{0.011} & 0.997 \std{0.007} & 1.0 \std{0.0} & 0.905 \std{0.004} & 0.084 \std{0.054} & 0.966 \std{0.003} & 0.254 \std{0.147} & 0.014 \std{0.011} \\
JNK3 & 0.654 \std{0.062} & 0.0 \std{0.0} & 1.0 \std{0.0} & 0.964 \std{0.025} & 0.885 \std{0.009} & 0.0 \std{0.0} & 0.948 \std{0.01} & 0.0 \std{0.0} & 0.0 \std{0.0} \\
GSK3 & 0.594 \std{0.065} & 0.002 \std{0.004} & 1.0 \std{0.0} & 0.966 \std{0.018} & 0.888 \std{0.007} & 0.016 \std{0.035} & 0.95 \std{0.008} & 0.055 \std{0.123} & 0.002 \std{0.004} \\
\midrule
BBB,QED & 0.766 \std{0.022} & 0.574 \std{0.064} & 0.978 \std{0.013} & 0.82 \std{0.043} & 0.846 \std{0.027} & 0.662 \std{0.043} & 0.878 \std{0.025} & 0.762 \std{0.02} & 0.424 \std{0.038} \\
HIA,QED & 0.796 \std{0.022} & 0.586 \std{0.036} & 0.996 \std{0.008} & 0.765 \std{0.059} & 0.815 \std{0.021} & 0.683 \std{0.027} & 0.853 \std{0.03} & 0.763 \std{0.019} & 0.396 \std{0.068} \\
QED,SA & 0.856 \std{0.028} & 0.634 \std{0.076} & 0.978 \std{0.025} & 0.615 \std{0.033} & 0.755 \std{0.018} & 0.736 \std{0.054} & 0.768 \std{0.019} & 0.751 \std{0.024} & 0.306 \std{0.059} \\
\midrule
LogP,-1 & 0.96 \std{0.024} & 0.36 \std{0.062} & 1.0 \std{0.0} & 0.347 \std{0.038} & 0.746 \std{0.031} & 0.586 \std{0.056} & 0.636 \std{0.03} & 0.61 \std{0.03} & 0.132 \std{0.019} \\
LogP,-3 & 0.92 \std{0.025} & 0.01 \std{0.01} & 0.979 \std{0.012} & 0.451 \std{0.061} & 0.776 \std{0.03} & 0.074 \std{0.07} & 0.699 \std{0.043} & 0.177 \std{0.163} & 0.01 \std{0.01} \\
LogP,1 & 0.946 \std{0.033} & 0.774 \std{0.059} & 0.988 \std{0.016} & 0.358 \std{0.017} & 0.785 \std{0.021} & 0.855 \std{0.044} & 0.652 \std{0.014} & 0.747 \std{0.015} & 0.218 \std{0.026} \\
LogP,3 & 0.87 \std{0.025} & 0.144 \std{0.019} & 1.0 \std{0.0} & 0.528 \std{0.039} & 0.692 \std{0.024} & 0.353 \std{0.028} & 0.715 \std{0.025} & 0.502 \std{0.025} & 0.134 \std{0.021} \\
LogP,5 & 0.796 \std{0.031} & 0.098 \std{0.027} & 0.98 \std{0.014} & 0.742 \std{0.06} & 0.804 \std{0.022} & 0.277 \std{0.036} & 0.836 \std{0.03} & 0.481 \std{0.039} & 0.098 \std{0.027} \\
\bottomrule
\end{tabular}
}
\end{table}

\begin{table}
\caption{Detailed results for LLaMA3-70B}
\label{tab:results_llama3-70b}
\resizebox{\textwidth}{!}{
\begin{tabular}{lccccccccc}
\toprule
Task & Validity & SR & Novelty & Uniqueness & Diversity & CC & DC & Creativity & Fully Creative \\
\midrule
BBB & 0.99 \std{0.007} & 0.98 \std{0.012} & 0.975 \std{0.025} & 0.374 \std{0.049} & 0.474 \std{0.045} & 0.985 \std{0.009} & 0.556 \std{0.038} & 0.74 \std{0.025} & 0.35 \std{0.035} \\
HIA & 0.996 \std{0.009} & 0.982 \std{0.025} & 1.0 \std{0.0} & 0.233 \std{0.045} & 0.307 \std{0.074} & 0.989 \std{0.015} & 0.413 \std{0.061} & 0.637 \std{0.046} & 0.22 \std{0.037} \\
\midrule
QED & 0.988 \std{0.016} & 0.946 \std{0.025} & 0.99 \std{0.021} & 0.178 \std{0.043} & 0.223 \std{0.041} & 0.967 \std{0.02} & 0.338 \std{0.049} & 0.571 \std{0.043} & 0.134 \std{0.039} \\
SA & 0.998 \std{0.004} & 0.994 \std{0.009} & 0.983 \std{0.037} & 0.124 \std{0.034} & 0.181 \std{0.045} & 0.996 \std{0.005} & 0.279 \std{0.049} & 0.526 \std{0.046} & 0.118 \std{0.033} \\
\midrule
DRD2 & 0.924 \std{0.03} & 0.024 \std{0.025} & 1.0 \std{0.0} & 0.564 \std{0.051} & 0.655 \std{0.04} & 0.114 \std{0.109} & 0.717 \std{0.036} & 0.224 \std{0.207} & 0.022 \std{0.023} \\
JNK3 & 0.92 \std{0.016} & 0.006 \std{0.009} & 0.984 \std{0.016} & 0.543 \std{0.026} & 0.589 \std{0.024} & 0.046 \std{0.065} & 0.68 \std{0.02} & 0.113 \std{0.155} & 0.006 \std{0.009} \\
GSK3 & 0.88 \std{0.031} & 0.012 \std{0.013} & 0.99 \std{0.013} & 0.49 \std{0.056} & 0.54 \std{0.058} & 0.078 \std{0.075} & 0.639 \std{0.047} & 0.176 \std{0.162} & 0.01 \std{0.01} \\
\midrule
BBB,QED & 0.978 \std{0.008} & 0.93 \std{0.01} & 0.948 \std{0.05} & 0.329 \std{0.026} & 0.396 \std{0.027} & 0.954 \std{0.005} & 0.498 \std{0.03} & 0.689 \std{0.021} & 0.26 \std{0.042} \\
HIA,QED & 0.99 \std{0.01} & 0.94 \std{0.016} & 0.975 \std{0.037} & 0.21 \std{0.038} & 0.273 \std{0.043} & 0.965 \std{0.011} & 0.381 \std{0.041} & 0.606 \std{0.034} & 0.152 \std{0.026} \\
QED,SA & 0.996 \std{0.005} & 0.97 \std{0.017} & 0.976 \std{0.033} & 0.163 \std{0.026} & 0.177 \std{0.033} & 0.983 \std{0.01} & 0.303 \std{0.035} & 0.545 \std{0.032} & 0.132 \std{0.028} \\
\midrule
LogP,-1 & 0.978 \std{0.022} & 0.378 \std{0.028} & 1.0 \std{0.0} & 0.288 \std{0.039} & 0.647 \std{0.046} & 0.608 \std{0.027} & 0.57 \std{0.032} & 0.588 \std{0.022} & 0.148 \std{0.025} \\
LogP,-3 & 0.984 \std{0.015} & 0.012 \std{0.011} & 0.995 \std{0.01} & 0.405 \std{0.038} & 0.745 \std{0.023} & 0.094 \std{0.061} & 0.669 \std{0.027} & 0.224 \std{0.131} & 0.012 \std{0.011} \\
LogP,1 & 0.994 \std{0.009} & 0.614 \std{0.039} & 0.973 \std{0.012} & 0.457 \std{0.035} & 0.769 \std{0.009} & 0.781 \std{0.028} & 0.699 \std{0.021} & 0.739 \std{0.008} & 0.276 \std{0.029} \\
LogP,3 & 0.982 \std{0.013} & 0.582 \std{0.045} & 0.995 \std{0.007} & 0.831 \std{0.052} & 0.785 \std{0.006} & 0.755 \std{0.03} & 0.866 \std{0.019} & 0.808 \std{0.021} & 0.492 \std{0.07} \\
LogP,5 & 0.986 \std{0.015} & 0.306 \std{0.048} & 1.0 \std{0.0} & 0.781 \std{0.044} & 0.547 \std{0.033} & 0.548 \std{0.048} & 0.753 \std{0.023} & 0.642 \std{0.028} & 0.204 \std{0.048} \\
\bottomrule
\end{tabular}
}
\end{table}

\begin{table}
\caption{Detailed results for LLaMA3.1-70B}
\label{tab:results_llama3.1-70b}
\resizebox{\textwidth}{!}{
\begin{tabular}{lccccccccc}
\toprule
Task & Validity & SR & Novelty & Uniqueness & Diversity & CC & DC & Creativity & Fully Creative \\
\midrule
BBB & 0.992 \std{0.008} & 0.97 \std{0.019} & 0.985 \std{0.021} & 0.24 \std{0.046} & 0.327 \std{0.062} & 0.981 \std{0.011} & 0.424 \std{0.052} & 0.644 \std{0.041} & 0.212 \std{0.04} \\
HIA & 0.998 \std{0.004} & 0.998 \std{0.004} & 1.0 \std{0.0} & 0.068 \std{0.024} & 0.081 \std{0.036} & 0.998 \std{0.004} & 0.174 \std{0.047} & 0.413 \std{0.059} & 0.068 \std{0.024} \\
\midrule
QED & 0.966 \std{0.011} & 0.918 \std{0.019} & 0.988 \std{0.028} & 0.221 \std{0.037} & 0.303 \std{0.05} & 0.942 \std{0.011} & 0.404 \std{0.047} & 0.616 \std{0.034} & 0.166 \std{0.033} \\
SA & 1.0 \std{0.0} & 1.0 \std{0.0} & 1.0 \std{0.0} & 0.058 \std{0.031} & 0.061 \std{0.034} & 1.0 \std{0.0} & 0.148 \std{0.054} & 0.38 \std{0.069} & 0.058 \std{0.031} \\
\midrule
DRD2 & 0.878 \std{0.029} & 0.036 \std{0.029} & 0.987 \std{0.009} & 0.861 \std{0.025} & 0.847 \std{0.004} & 0.167 \std{0.069} & 0.896 \std{0.009} & 0.38 \std{0.077} & 0.028 \std{0.02} \\
JNK3 & 0.94 \std{0.025} & 0.0 \std{0.0} & 0.972 \std{0.034} & 0.318 \std{0.059} & 0.394 \std{0.057} & 0.0 \std{0.0} & 0.494 \std{0.056} & 0.0 \std{0.0} & 0.0 \std{0.0} \\
GSK3 & 0.894 \std{0.025} & 0.004 \std{0.005} & 1.0 \std{0.0} & 0.403 \std{0.058} & 0.462 \std{0.054} & 0.038 \std{0.052} & 0.57 \std{0.049} & 0.091 \std{0.125} & 0.004 \std{0.005} \\
\midrule
BBB,QED & 0.976 \std{0.025} & 0.894 \std{0.035} & 0.969 \std{0.015} & 0.402 \std{0.047} & 0.53 \std{0.049} & 0.934 \std{0.03} & 0.591 \std{0.041} & 0.742 \std{0.027} & 0.3 \std{0.047} \\
HIA,QED & 0.986 \std{0.015} & 0.924 \std{0.027} & 0.969 \std{0.029} & 0.18 \std{0.046} & 0.276 \std{0.064} & 0.955 \std{0.021} & 0.361 \std{0.056} & 0.586 \std{0.047} & 0.138 \std{0.033} \\
QED,SA & 0.978 \std{0.016} & 0.938 \std{0.026} & 0.965 \std{0.032} & 0.182 \std{0.017} & 0.239 \std{0.031} & 0.958 \std{0.02} & 0.347 \std{0.028} & 0.576 \std{0.019} & 0.138 \std{0.026} \\
\midrule
LogP,-1 & 1.0 \std{0.0} & 0.044 \std{0.019} & 1.0 \std{0.0} & 0.144 \std{0.015} & 0.366 \std{0.015} & 0.203 \std{0.059} & 0.375 \std{0.015} & 0.274 \std{0.049} & 0.032 \std{0.019} \\
LogP,-3 & 0.986 \std{0.011} & 0.002 \std{0.004} & 1.0 \std{0.0} & 0.231 \std{0.048} & 0.438 \std{0.043} & 0.02 \std{0.044} & 0.465 \std{0.044} & 0.042 \std{0.094} & 0.002 \std{0.004} \\
LogP,1 & 0.994 \std{0.005} & 0.934 \std{0.011} & 0.993 \std{0.015} & 0.294 \std{0.027} & 0.698 \std{0.015} & 0.964 \std{0.006} & 0.588 \std{0.019} & 0.753 \std{0.011} & 0.24 \std{0.016} \\
LogP,3 & 0.988 \std{0.013} & 0.684 \std{0.059} & 0.981 \std{0.006} & 0.634 \std{0.03} & 0.807 \std{0.005} & 0.821 \std{0.036} & 0.794 \std{0.013} & 0.807 \std{0.019} & 0.362 \std{0.054} \\
LogP,5 & 0.998 \std{0.004} & 0.142 \std{0.028} & 1.0 \std{0.0} & 0.549 \std{0.02} & 0.465 \std{0.014} & 0.375 \std{0.035} & 0.634 \std{0.012} & 0.487 \std{0.021} & 0.092 \std{0.016} \\
\bottomrule
\end{tabular}
}
\end{table}

\begin{table}
\caption{Detailed results for GPT-3.5}
\label{tab:results_gpt-3.5-turbo}
\resizebox{\textwidth}{!}{
\begin{tabular}{lccccccccc}
\toprule
Task & Validity & SR & Novelty & Uniqueness & Diversity & CC & DC & Creativity & Fully Creative \\
\midrule
BBB & 0.988 \std{0.008} & 0.888 \std{0.016} & 0.989 \std{0.006} & 0.763 \std{0.014} & 0.882 \std{0.01} & 0.937 \std{0.011} & 0.873 \std{0.007} & 0.904 \std{0.005} & 0.656 \std{0.015} \\
HIA & 0.986 \std{0.017} & 0.97 \std{0.014} & 0.995 \std{0.011} & 0.42 \std{0.027} & 0.677 \std{0.026} & 0.978 \std{0.014} & 0.656 \std{0.019} & 0.801 \std{0.01} & 0.396 \std{0.029} \\
\midrule
QED & 0.97 \std{0.016} & 0.636 \std{0.03} & 0.963 \std{0.021} & 0.892 \std{0.05} & 0.882 \std{0.008} & 0.785 \std{0.017} & 0.912 \std{0.019} & 0.846 \std{0.014} & 0.566 \std{0.038} \\
SA & 0.998 \std{0.004} & 0.996 \std{0.009} & 1.0 \std{0.0} & 0.423 \std{0.033} & 0.791 \std{0.015} & 0.997 \std{0.007} & 0.694 \std{0.022} & 0.832 \std{0.011} & 0.42 \std{0.028} \\
\midrule
DRD2 & 0.978 \std{0.015} & 0.15 \std{0.019} & 0.981 \std{0.014} & 0.957 \std{0.018} & 0.879 \std{0.006} & 0.382 \std{0.023} & 0.938 \std{0.01} & 0.598 \std{0.019} & 0.15 \std{0.019} \\
JNK3 & 0.976 \std{0.013} & 0.006 \std{0.005} & 0.972 \std{0.012} & 0.951 \std{0.009} & 0.86 \std{0.005} & 0.059 \std{0.054} & 0.926 \std{0.005} & 0.181 \std{0.165} & 0.006 \std{0.005} \\
GSK3 & 0.952 \std{0.016} & 0.032 \std{0.019} & 0.976 \std{0.024} & 0.965 \std{0.012} & 0.875 \std{0.004} & 0.168 \std{0.054} & 0.937 \std{0.008} & 0.392 \std{0.066} & 0.032 \std{0.019} \\
\midrule
BBB,QED & 0.974 \std{0.011} & 0.642 \std{0.036} & 0.967 \std{0.011} & 0.941 \std{0.024} & 0.882 \std{0.009} & 0.79 \std{0.018} & 0.929 \std{0.005} & 0.857 \std{0.011} & 0.576 \std{0.047} \\
HIA,QED & 0.978 \std{0.018} & 0.416 \std{0.065} & 0.985 \std{0.015} & 0.671 \std{0.034} & 0.836 \std{0.011} & 0.636 \std{0.048} & 0.82 \std{0.017} & 0.722 \std{0.034} & 0.388 \std{0.064} \\
QED,SA & 0.964 \std{0.017} & 0.704 \std{0.042} & 0.98 \std{0.01} & 0.929 \std{0.025} & 0.875 \std{0.006} & 0.824 \std{0.03} & 0.927 \std{0.012} & 0.874 \std{0.016} & 0.66 \std{0.034} \\
\midrule
LogP,-1 & 1.0 \std{0.0} & 0.776 \std{0.013} & 1.0 \std{0.0} & 0.182 \std{0.029} & 0.357 \std{0.039} & 0.881 \std{0.008} & 0.401 \std{0.033} & 0.594 \std{0.026} & 0.036 \std{0.033} \\
LogP,-3 & 1.0 \std{0.0} & 0.006 \std{0.005} & 1.0 \std{0.0} & 0.366 \std{0.021} & 0.672 \std{0.018} & 0.06 \std{0.055} & 0.626 \std{0.015} & 0.151 \std{0.138} & 0.006 \std{0.005} \\
LogP,1 & 1.0 \std{0.0} & 0.992 \std{0.008} & 1.0 \std{0.0} & 0.086 \std{0.015} & 0.189 \std{0.027} & 0.996 \std{0.004} & 0.252 \std{0.021} & 0.501 \std{0.021} & 0.078 \std{0.016} \\
LogP,3 & 1.0 \std{0.0} & 0.672 \std{0.044} & 0.993 \std{0.016} & 0.458 \std{0.068} & 0.71 \std{0.04} & 0.82 \std{0.027} & 0.685 \std{0.041} & 0.748 \std{0.017} & 0.212 \std{0.038} \\
LogP,5 & 0.998 \std{0.004} & 0.314 \std{0.047} & 0.997 \std{0.006} & 0.798 \std{0.04} & 0.761 \std{0.012} & 0.558 \std{0.043} & 0.846 \std{0.017} & 0.687 \std{0.026} & 0.228 \std{0.038} \\
\bottomrule
\end{tabular}
}
\end{table}

\begin{table}
\caption{Detailed results for GPT-4o-mini}
\label{tab:results_gpt-4o-mini}
\resizebox{\textwidth}{!}{
\begin{tabular}{lccccccccc}
\toprule
Task & Validity & SR & Novelty & Uniqueness & Diversity & CC & DC & Creativity & Fully Creative \\
\midrule
BBB & 0.938 \std{0.028} & 0.812 \std{0.052} & 0.99 \std{0.009} & 0.68 \std{0.052} & 0.819 \std{0.017} & 0.872 \std{0.032} & 0.82 \std{0.028} & 0.845 \std{0.018} & 0.516 \std{0.037} \\
HIA & 0.978 \std{0.008} & 0.942 \std{0.011} & 0.971 \std{0.02} & 0.342 \std{0.046} & 0.656 \std{0.029} & 0.96 \std{0.008} & 0.6 \std{0.029} & 0.759 \std{0.018} & 0.29 \std{0.035} \\
\midrule
QED & 0.912 \std{0.029} & 0.552 \std{0.07} & 0.992 \std{0.019} & 0.656 \std{0.09} & 0.801 \std{0.018} & 0.709 \std{0.052} & 0.804 \std{0.039} & 0.754 \std{0.039} & 0.388 \std{0.074} \\
SA & 0.992 \std{0.008} & 0.992 \std{0.008} & 0.978 \std{0.02} & 0.268 \std{0.029} & 0.547 \std{0.041} & 0.992 \std{0.008} & 0.522 \std{0.026} & 0.72 \std{0.021} & 0.26 \std{0.029} \\
\midrule
DRD2 & 0.888 \std{0.036} & 0.0 \std{0.0} & 0.997 \std{0.006} & 0.796 \std{0.039} & 0.814 \std{0.022} & 0.0 \std{0.0} & 0.865 \std{0.021} & 0.0 \std{0.0} & 0.0 \std{0.0} \\
JNK3 & 0.844 \std{0.034} & 0.0 \std{0.0} & 0.995 \std{0.007} & 0.873 \std{0.047} & 0.844 \std{0.014} & 0.0 \std{0.0} & 0.901 \std{0.023} & 0.0 \std{0.0} & 0.0 \std{0.0} \\
GSK3 & 0.796 \std{0.051} & 0.0 \std{0.0} & 0.983 \std{0.015} & 0.903 \std{0.038} & 0.849 \std{0.012} & 0.0 \std{0.0} & 0.91 \std{0.015} & 0.0 \std{0.0} & 0.0 \std{0.0} \\
\midrule
BBB,QED & 0.94 \std{0.023} & 0.758 \std{0.013} & 0.99 \std{0.01} & 0.618 \std{0.053} & 0.775 \std{0.016} & 0.844 \std{0.013} & 0.779 \std{0.026} & 0.811 \std{0.019} & 0.402 \std{0.055} \\
HIA,QED & 0.97 \std{0.023} & 0.62 \std{0.025} & 0.97 \std{0.02} & 0.484 \std{0.033} & 0.765 \std{0.01} & 0.775 \std{0.024} & 0.711 \std{0.02} & 0.742 \std{0.016} & 0.31 \std{0.026} \\
QED,SA & 0.95 \std{0.027} & 0.606 \std{0.023} & 0.981 \std{0.002} & 0.561 \std{0.044} & 0.764 \std{0.005} & 0.759 \std{0.022} & 0.749 \std{0.021} & 0.753 \std{0.015} & 0.39 \std{0.026} \\
\midrule
LogP,-1 & 0.982 \std{0.015} & 0.466 \std{0.027} & 0.995 \std{0.011} & 0.416 \std{0.022} & 0.708 \std{0.008} & 0.676 \std{0.017} & 0.664 \std{0.013} & 0.67 \std{0.013} & 0.164 \std{0.031} \\
LogP,-3 & 0.988 \std{0.011} & 0.014 \std{0.013} & 0.979 \std{0.014} & 0.579 \std{0.058} & 0.748 \std{0.018} & 0.091 \std{0.084} & 0.751 \std{0.03} & 0.202 \std{0.185} & 0.014 \std{0.013} \\
LogP,1 & 0.998 \std{0.004} & 0.57 \std{0.025} & 1.0 \std{0.0} & 0.597 \std{0.024} & 0.773 \std{0.009} & 0.754 \std{0.016} & 0.773 \std{0.012} & 0.763 \std{0.01} & 0.318 \std{0.028} \\
LogP,3 & 0.98 \std{0.021} & 0.442 \std{0.069} & 0.986 \std{0.011} & 0.725 \std{0.051} & 0.792 \std{0.01} & 0.656 \std{0.055} & 0.827 \std{0.024} & 0.736 \std{0.032} & 0.322 \std{0.061} \\
LogP,5 & 0.952 \std{0.015} & 0.246 \std{0.017} & 0.983 \std{0.007} & 0.862 \std{0.025} & 0.828 \std{0.009} & 0.483 \std{0.017} & 0.888 \std{0.011} & 0.655 \std{0.01} & 0.216 \std{0.023} \\
\bottomrule
\end{tabular}
}
\end{table}

\begin{table}
\caption{Detailed results for DeepSeek-V3}
\label{tab:results_deepseek-chat-v3-0324}
\resizebox{\textwidth}{!}{
\begin{tabular}{lccccccccc}
\toprule
Task & Validity & SR & Novelty & Uniqueness & Diversity & CC & DC & Creativity & Fully Creative \\
\midrule
BBB & 0.986 \std{0.011} & 0.982 \std{0.011} & 0.949 \std{0.008} & 0.205 \std{0.032} & 0.695 \std{0.034} & 0.984 \std{0.011} & 0.513 \std{0.036} & 0.71 \std{0.026} & 0.188 \std{0.03} \\
HIA & 1.0 \std{0.0} & 0.996 \std{0.009} & 0.962 \std{0.052} & 0.098 \std{0.008} & 0.258 \std{0.049} & 0.998 \std{0.004} & 0.289 \std{0.022} & 0.536 \std{0.019} & 0.09 \std{0.007} \\
\midrule
QED & 1.0 \std{0.0} & 0.402 \std{0.045} & 0.919 \std{0.017} & 0.34 \std{0.047} & 0.611 \std{0.045} & 0.633 \std{0.036} & 0.575 \std{0.039} & 0.603 \std{0.035} & 0.244 \std{0.025} \\
SA & 1.0 \std{0.0} & 0.998 \std{0.004} & 0.98 \std{0.045} & 0.1 \std{0.012} & 0.523 \std{0.015} & 0.999 \std{0.002} & 0.371 \std{0.017} & 0.608 \std{0.014} & 0.096 \std{0.015} \\
\midrule
DRD2 & 0.99 \std{0.01} & 0.134 \std{0.03} & 0.995 \std{0.011} & 0.434 \std{0.042} & 0.682 \std{0.033} & 0.362 \std{0.039} & 0.665 \std{0.033} & 0.491 \std{0.031} & 0.084 \std{0.019} \\
JNK3 & 0.992 \std{0.008} & 0.004 \std{0.009} & 0.965 \std{0.01} & 0.929 \std{0.031} & 0.868 \std{0.003} & 0.028 \std{0.063} & 0.92 \std{0.013} & 0.072 \std{0.161} & 0.004 \std{0.009} \\
GSK3 & 0.992 \std{0.008} & 0.022 \std{0.018} & 0.944 \std{0.02} & 0.75 \std{0.035} & 0.852 \std{0.006} & 0.129 \std{0.081} & 0.845 \std{0.014} & 0.294 \std{0.169} & 0.022 \std{0.018} \\
\midrule
BBB,QED & 0.99 \std{0.007} & 0.398 \std{0.062} & 0.892 \std{0.051} & 0.303 \std{0.027} & 0.717 \std{0.028} & 0.626 \std{0.049} & 0.578 \std{0.024} & 0.601 \std{0.028} & 0.196 \std{0.029} \\
HIA,QED & 1.0 \std{0.0} & 0.202 \std{0.029} & 0.908 \std{0.054} & 0.206 \std{0.026} & 0.376 \std{0.027} & 0.448 \std{0.032} & 0.412 \std{0.019} & 0.43 \std{0.021} & 0.134 \std{0.015} \\
QED,SA & 1.0 \std{0.0} & 0.67 \std{0.033} & 0.906 \std{0.037} & 0.366 \std{0.021} & 0.72 \std{0.02} & 0.818 \std{0.02} & 0.62 \std{0.024} & 0.712 \std{0.014} & 0.266 \std{0.034} \\
\midrule
LogP,-1 & 1.0 \std{0.0} & 0.886 \std{0.03} & 1.0 \std{0.0} & 0.158 \std{0.037} & 0.611 \std{0.022} & 0.941 \std{0.016} & 0.457 \std{0.042} & 0.655 \std{0.033} & 0.1 \std{0.032} \\
LogP,-3 & 0.996 \std{0.005} & 0.192 \std{0.052} & 1.0 \std{0.0} & 0.239 \std{0.042} & 0.576 \std{0.031} & 0.434 \std{0.06} & 0.515 \std{0.035} & 0.471 \std{0.032} & 0.086 \std{0.019} \\
LogP,1 & 1.0 \std{0.0} & 0.962 \std{0.011} & 0.971 \std{0.026} & 0.194 \std{0.023} & 0.622 \std{0.024} & 0.981 \std{0.005} & 0.489 \std{0.02} & 0.692 \std{0.015} & 0.158 \std{0.024} \\
LogP,3 & 1.0 \std{0.0} & 0.52 \std{0.06} & 1.0 \std{0.0} & 0.236 \std{0.038} & 0.583 \std{0.032} & 0.72 \std{0.042} & 0.515 \std{0.036} & 0.609 \std{0.033} & 0.154 \std{0.035} \\
LogP,5 & 1.0 \std{0.0} & 0.45 \std{0.069} & 0.957 \std{0.019} & 0.332 \std{0.059} & 0.515 \std{0.04} & 0.669 \std{0.05} & 0.545 \std{0.043} & 0.603 \std{0.021} & 0.082 \std{0.022} \\
\bottomrule
\end{tabular}
}
\end{table}

\section{Additional Scatter Visualizations for Numerical Constraint Task}

\label{app:logp_scatter}

\begin{figure*}[!ht]
  \centering
  \begin{subfigure}[b]{0.49\textwidth}
    \centering
    \includegraphics[width=\linewidth]{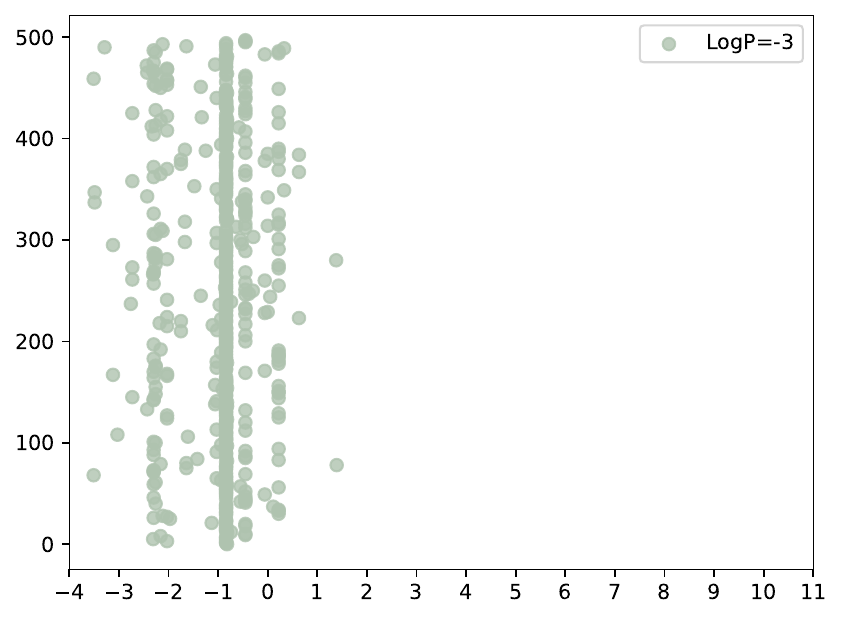}
    \caption{LogP=-3}
    \label{fig:LogP=-3}
  \end{subfigure}\hfill
  \begin{subfigure}[b]{0.49\textwidth}
    \centering
    \includegraphics[width=\linewidth]{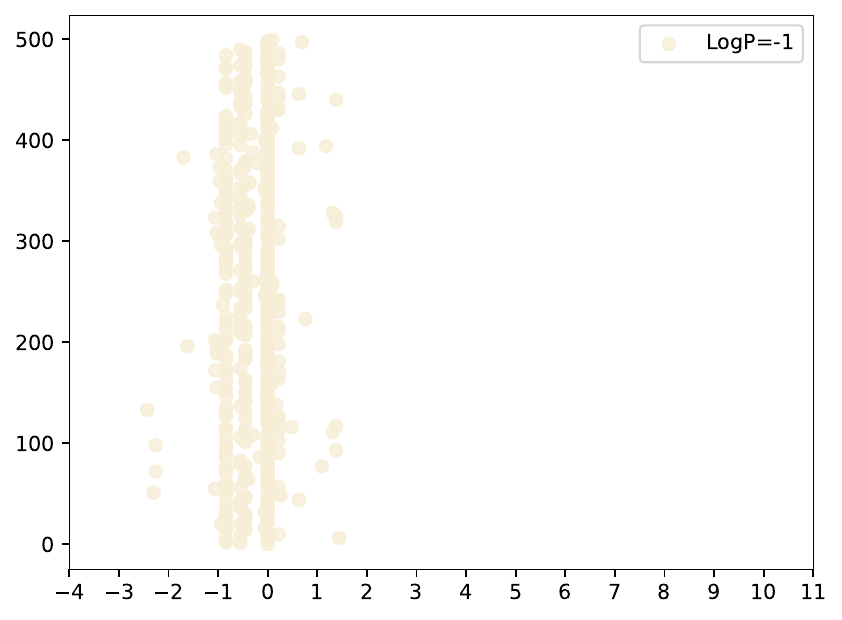}
    \caption{LogP=-1}
    \label{fig:LogP=-1}
  \end{subfigure}\hfill
  \begin{subfigure}[b]{0.49\textwidth}
    \centering
    \includegraphics[width=\linewidth]{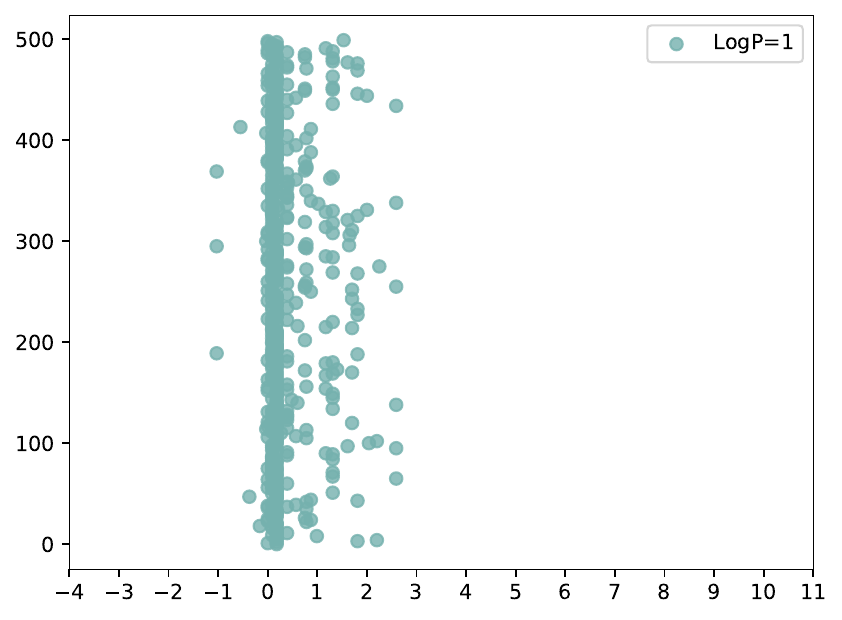}
    \caption{LogP=1}
    \label{fig:LogP=1}
  \end{subfigure}\hfill
  \begin{subfigure}[b]{0.49\textwidth}
    \centering
    \includegraphics[width=\linewidth]{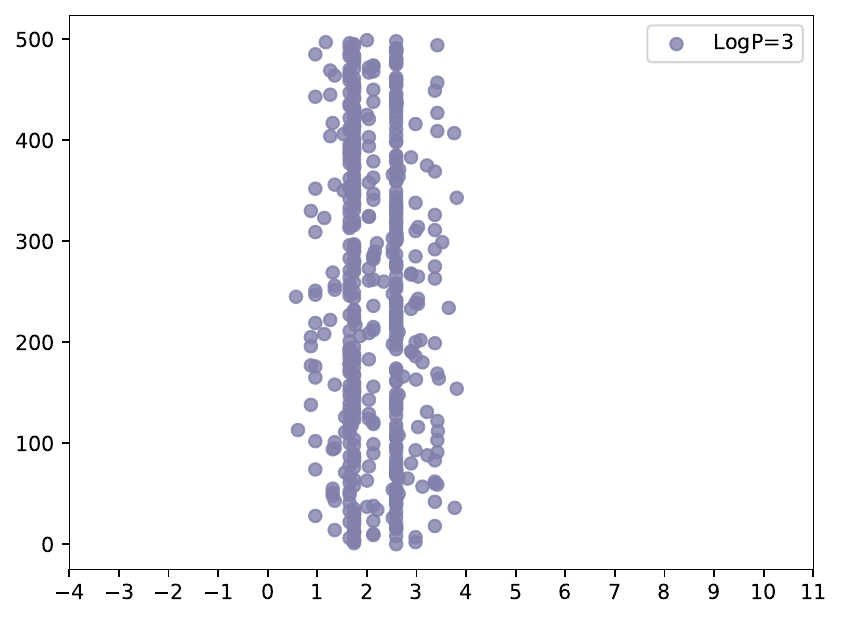}
    \caption{LogP=3}
    \label{fig:LogP=3}
  \end{subfigure}\hfill
  \begin{subfigure}[b]{0.49\textwidth}
    \centering
    \includegraphics[width=\linewidth]{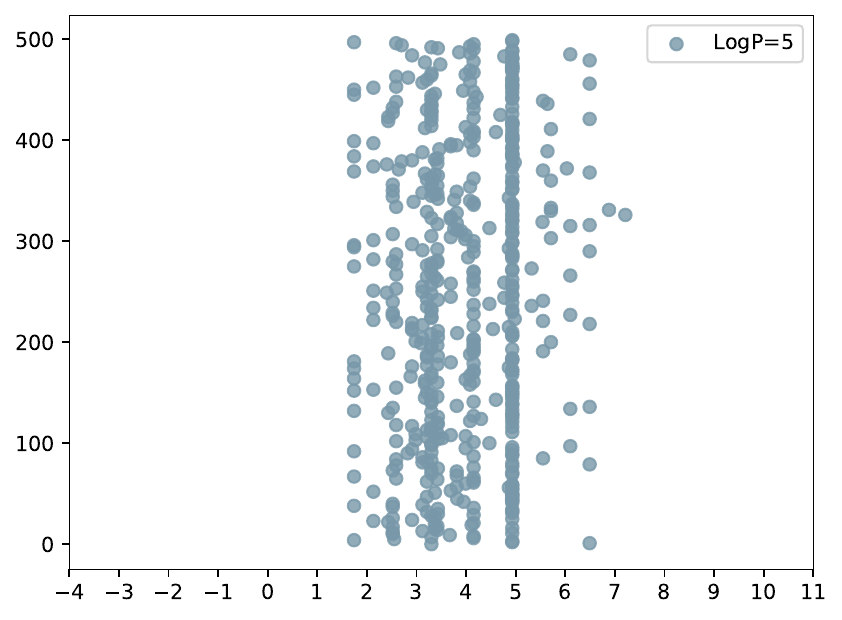}
    \caption{LogP=5}
    \label{fig:LogP=5}
  \end{subfigure}
  \caption{Scatter plots of molecular LogP distributions across different target conditions.}
  \label{fig:LogP_scatters}
\end{figure*}

\begin{figure}[!ht]
  \centering
    \centering
    \includegraphics[width=0.49\textwidth]{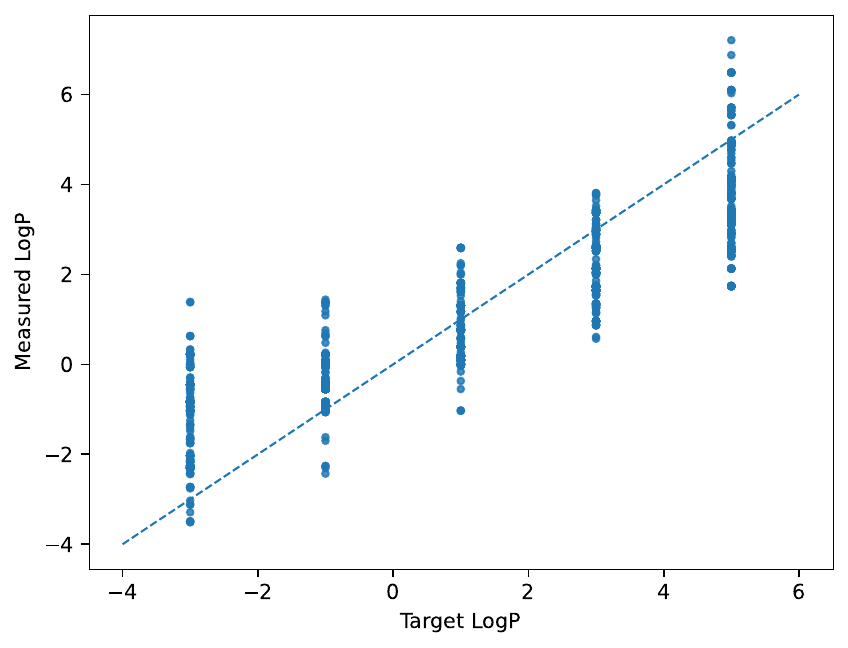}
  \caption{Conditional LogP distributions under discrete numerical constraints.}
  \label{fig:correlation_scatter}
\end{figure}

Figure~\ref{fig:LogP_scatters} and Figure~\ref{fig:correlation_scatter} show scatter plots of the LogP values of molecules generated by DeepSeek-V3 under different specified LogP targets. As the target LogP increases, the distribution of the generated molecules’ measured LogP shifts accordingly.

\section{Comparison of Overall Creativity}

\label{app:overall_creativity}

\begin{figure}[!ht]
  \centering
  \begin{subfigure}[b]{0.49\textwidth}
    \centering
    \includegraphics[width=\linewidth]{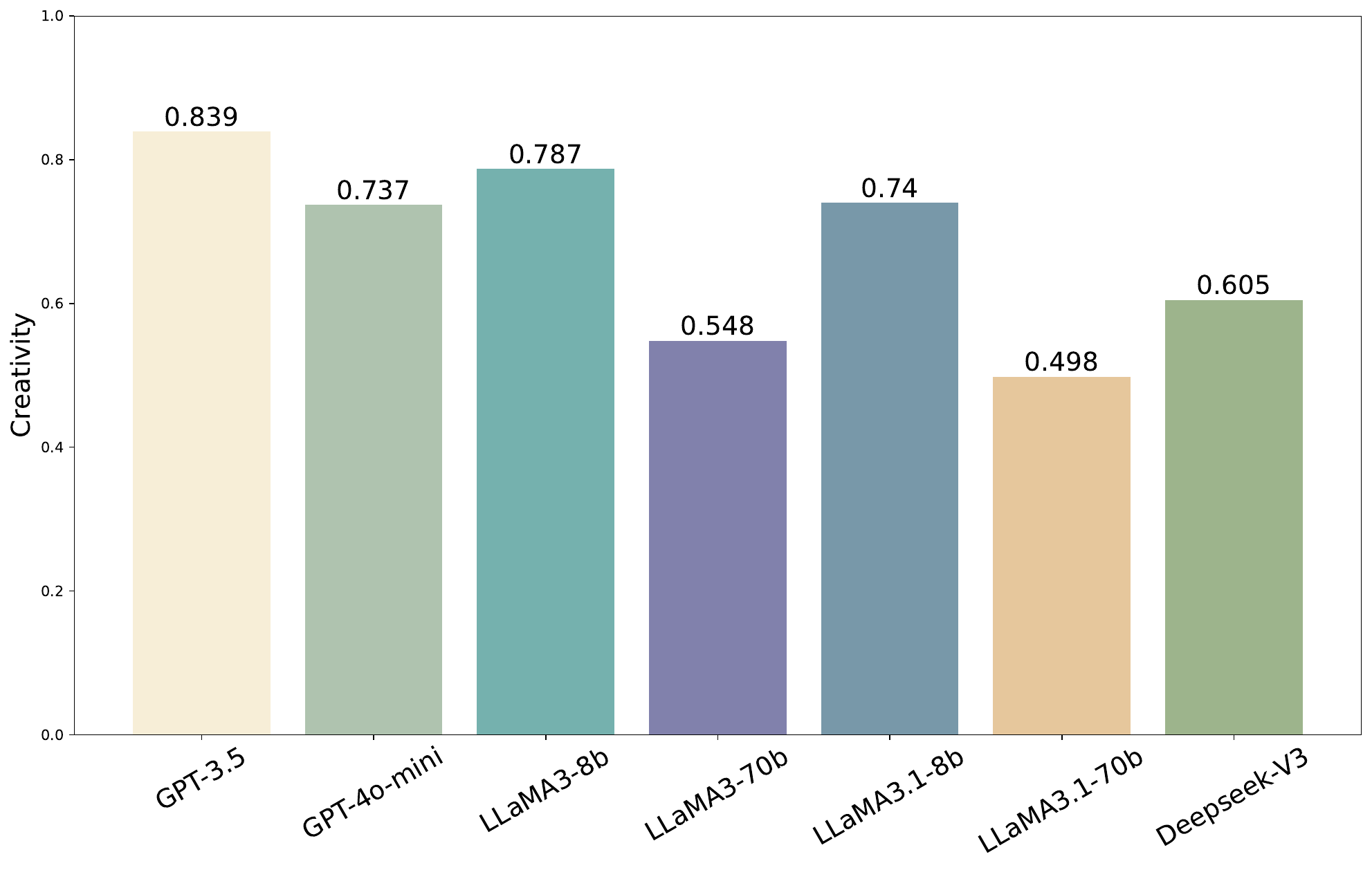}
    \caption{Physicochemical tasks}
    \label{fig:compare_creativity_physicochemical}
  \end{subfigure}
  \begin{subfigure}[b]{0.49\textwidth}
    \centering
    \includegraphics[width=\linewidth]{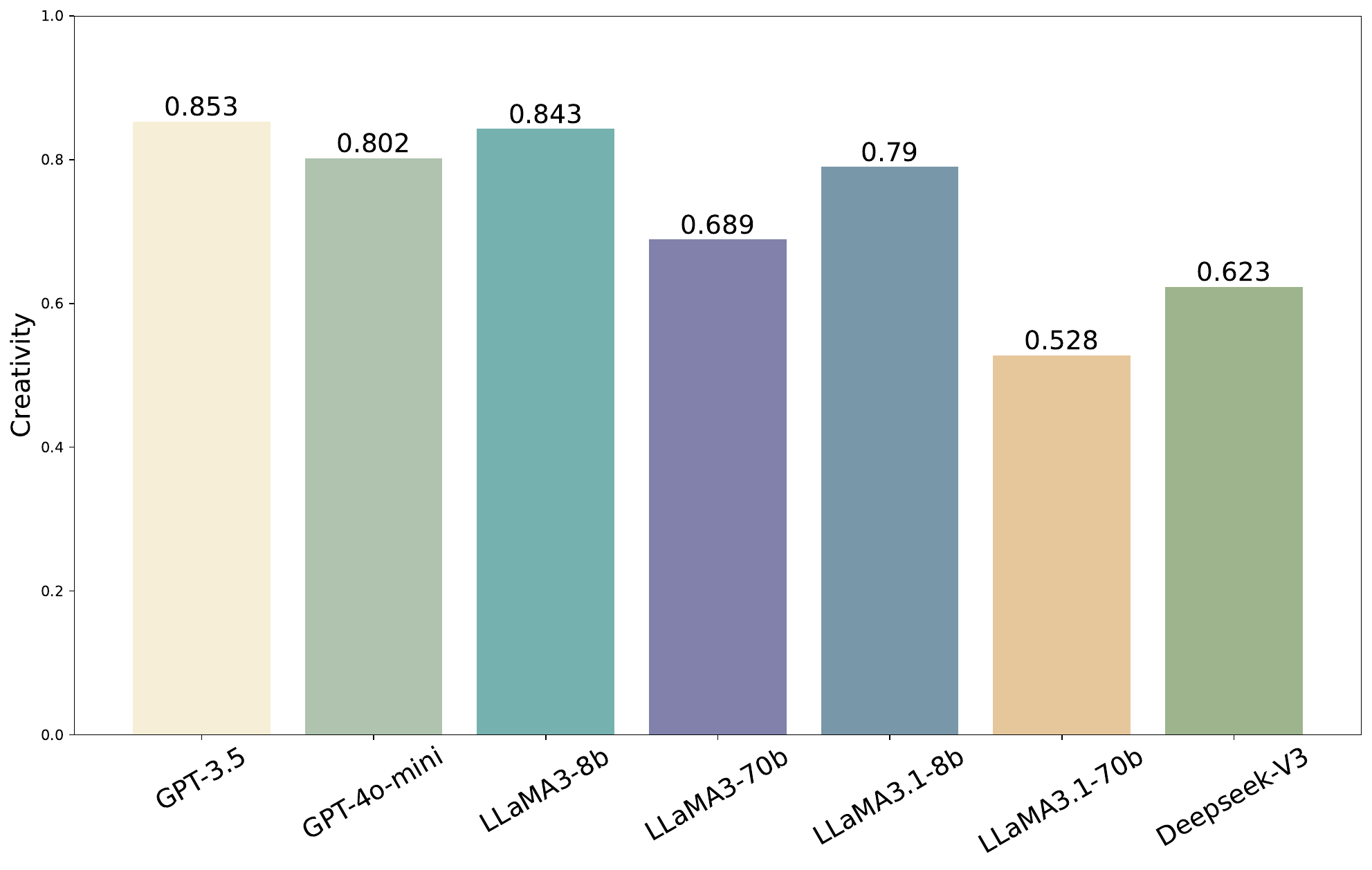}
    \caption{ADMET tasks}
    \label{fig:compare_creativity_admet}
  \end{subfigure}
  \begin{subfigure}[b]{0.49\textwidth}
    \centering
    \includegraphics[width=\linewidth]{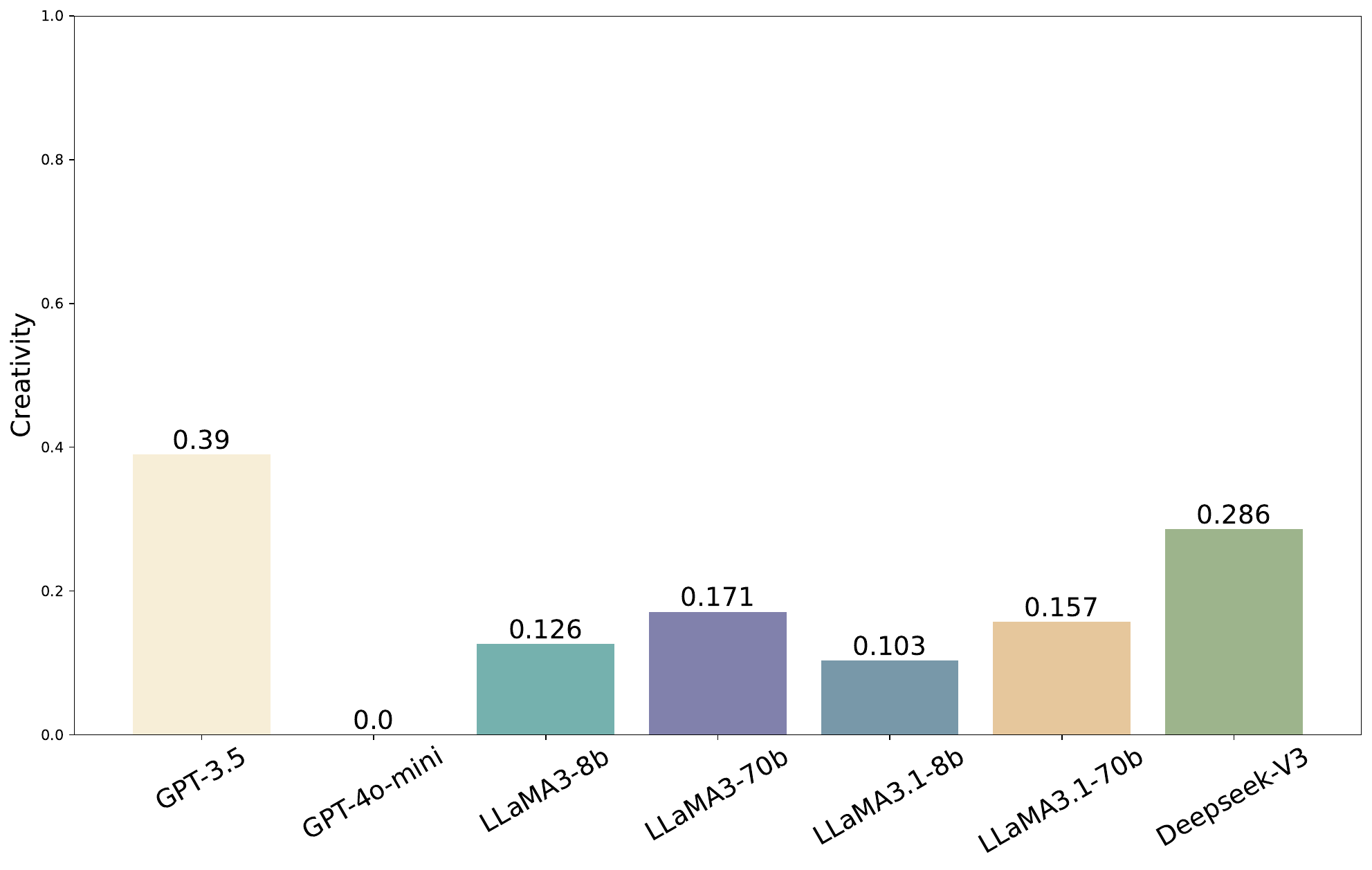}
    \caption{Activity tasks}
    \label{fig:compare_creativity_target}
  \end{subfigure}
  \begin{subfigure}[b]{0.49\textwidth}
    \centering
    \includegraphics[width=\linewidth]{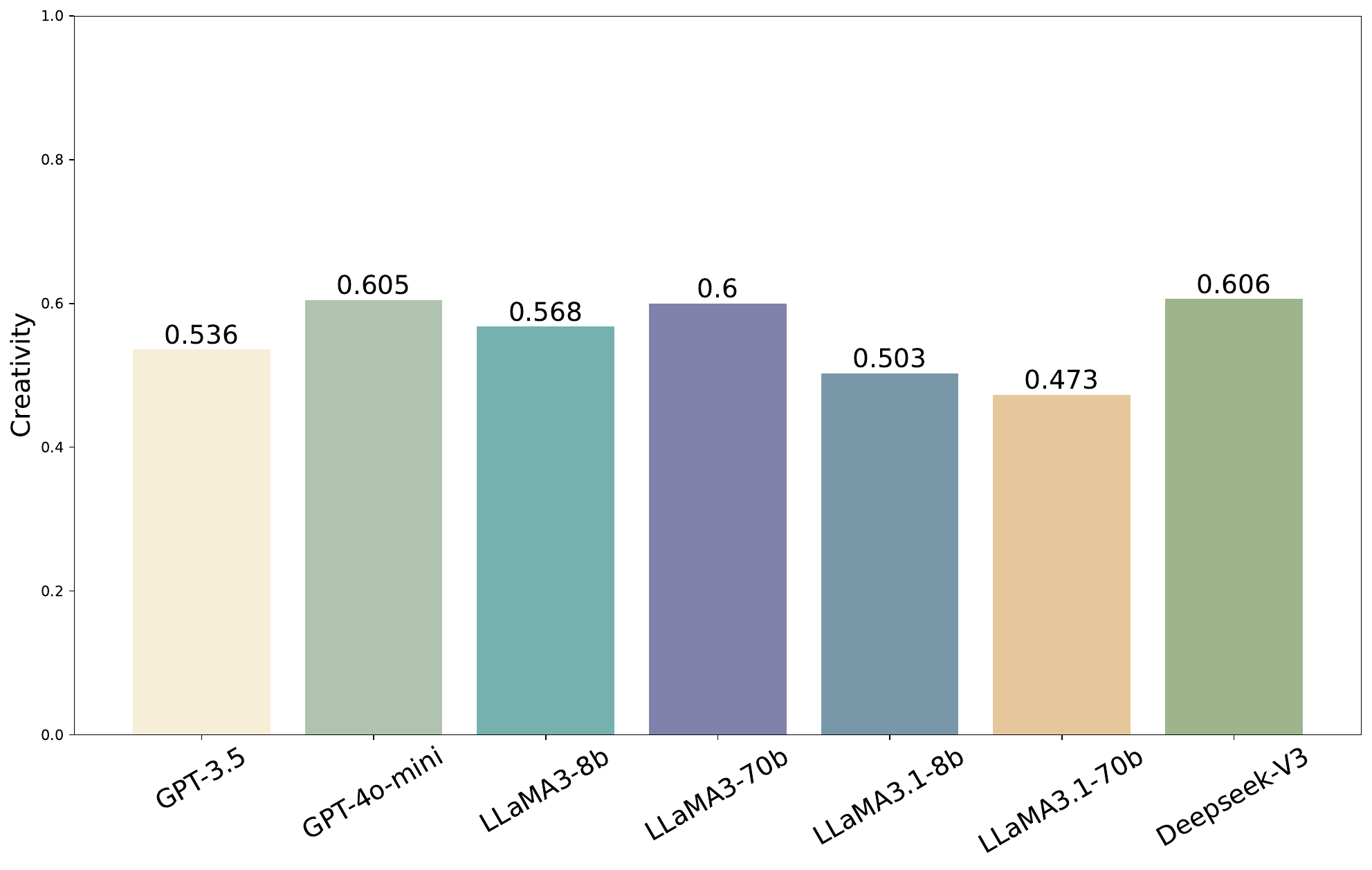}
    \caption{Numerical constraint tasks}
    \label{fig:compare_creativity_numerical}
  \end{subfigure}
  \caption{Average overall creativity of all models on physicochemical, ADMET, activity, and numerical constraint tasks.}
  \label{fig:compare_creativity}
\end{figure}

Figure~\ref{fig:compare_creativity} show the average overall creativity of all models across physicochemical, ADMET, activity, and numerical constraint tasks to facilitate comparison of creativity among models. Due to the special nature of numerical constraints, we evaluate this task type separately when comparing LLM creativity. The results show that GPT-3.5 achieves the highest Overall Creativity on physicochemical, ADMET and activity tasks. An exception is the numerical constraint task, where DeepSeek-V3 outperforms other models. We also find that smaller models, such as LLaMA3-8B and LLaMA3.1-8B, are not consistently the least creative across task types, indicating that model scale alone does not determine creativity in molecule generation.

\end{document}